\pdfoutput=1

\documentclass[11pt]{article}

\usepackage[final]{acl}

\usepackage{times}
\usepackage{latexsym}

\usepackage[T1]{fontenc}

\usepackage[utf8]{inputenc}

\usepackage{microtype}

\usepackage{inconsolata}

\usepackage{graphicx}
\usepackage{tabularx}
\usepackage{enumitem}
\usepackage{booktabs}
\usepackage{amsmath} 
\usepackage{amsfonts}
\usepackage{multirow}
\usepackage{soul}
\usepackage{array}
\usepackage{xspace}
\usepackage{xcolor}
\usepackage{soul}
\usepackage{makecell}

\definecolor{myblue}{HTML}{4273F5}
\definecolor{mypurple}{HTML}{9900FF}
\newcommand{\approach}{\textsc{Dr Genr\'e}\xspace}

\title{\approach: Reinforcement Learning from Decoupled LLM Feedback for Generic Text Rewriting}

\author{
Yufei Li$^{1}$\thanks{Work done as a research intern at Google DeepMind. Correspondence: Yufei Li (yli927@ucr.edu), John Nham (jnham@google.com), Ganesh Jawahar (ganayu@google.com).}
\xspace\xspace    
John Nham$^{2}$
\xspace\xspace
Ganesh Jawahar$^{2}$
\xspace\xspace
Lei Shu$^{2}$
\xspace\xspace
David Uthus$^{2}$ \\ \bf
Yun-Hsuan Sung$^{2}$
\xspace\xspace
Chengrun Yang$^{2}$
\xspace\xspace
Itai Rolnick$^{2}$
\xspace\xspace
Yi Qiao$^{3}$
\xspace\xspace
Cong Liu$^{1}$
\xspace\xspace\\
$^{1}$UC Riverside\xspace\xspace
$^{2}$Google DeepMind\xspace\xspace
$^{3}$Google Cloud\xspace\xspace
\\}

\begin{document}
\maketitle
\begin{abstract}
Generic text rewriting is a prevalent large language model (LLM) application that covers diverse real-world tasks, such as style transfer, fact correction, and email editing. These tasks vary in rewriting objectives (e.g., factual consistency vs. semantic preservation), making it challenging to develop a unified model that excels across all dimensions. Existing methods often specialize in either a single task or a specific objective, limiting their generalizability.
In this work, we introduce a generic model proficient in \emph{factuality}, \emph{stylistic}, and \emph{conversational} rewriting tasks.
To simulate real-world user rewrite requests, we construct a  conversational rewrite dataset, \textsc{ChatRewrite}, that presents ``natural''-sounding instructions, from raw emails using LLMs. Combined with other popular rewrite datasets, including \textsc{LongFact} for the factuality rewrite task and \textsc{RewriteLM} for the stylistic rewrite task, this forms a broad benchmark for training and evaluating generic rewrite models.
To align with task-specific objectives, we propose \approach, a \textbf{D}ecoupled-\textbf{r}eward learning framework for \textbf{Gen}eric \textbf{re}writing, that utilizes objective-oriented reward models with a task-specific weighting.
Evaluation shows that \approach delivers higher-quality rewrites across all targeted tasks, improving objectives including instruction following (\emph{agreement}), internal consistency (\emph{coherence}), and minimal unnecessary edits (\emph{conciseness}).
\end{abstract}


\section{Introduction}
Text rewriting is a fundamental NLP task with applications spanning style transfer~\cite{shu2024rewritelm}, summarization, and fact correction~\cite{wei2024long}. Existing models often specialize in a particular transformation type, such as paraphrasing~\cite{siddique2020unsupervised}, sentence fusion~\cite{mallinson2022edit5}, or focus on optimizing a specific objective, such as syntactic overlap with reference texts, during post-training~\cite{shen2017style,hu2017toward}. This specialization limits their applicability in real-world scenarios where diverse rewriting capabilities are required. In this work, we address the challenges of building a generic rewrite model capable of handling multiple rewriting tasks, including fact correction, style transfer, and conversational rewriting. 

\noindent\textbf{Versatile rewriting tasks.} \emph{Factuality rewrite} involves correcting content that contains factual errors in initial responses generated by a large language model (LLM) responding to fact-seeking prompts on open-ended topics~\cite{hu2023survey}. The rewrite model takes the prompt, initial response, and critique outputs (e.g., span-level checks from autoraters), and generates revised responses that correct non-factual claims, maintain global coherence (a challenge for standard posttraining methods, as shown in Table~\ref{tab:longfact_results}), and minimally edit the factual claims. 
\emph{Stylistic rewrite} focuses on transforming a source text into another style without introducing new information~\cite{shu2024rewritelm}, including formalization, paraphrasing, and summarization~\cite{li2018delete,zhang2020parallel}. 
While this task has been studied extensively in the in-context learning (ICL) setting~\cite{brown2020language,raffel2020exploring}, few-shot methods struggle to follow user-specified instructions accurately (as shown in Table~\ref{tab:rewritelm_results}).
Both factuality and stylistic rewrite expose limited user applicability. To mimic real-world user requests, we introduce a new rewriting task—\emph{conversational rewrite}—which addresses the need for modifying specific parts of a text based on user instructions that may lack detailed context. For instance, a user might say, ``This email is a bit dry; let's celebrate our success! Add some enthusiastic phrases like `We nailed it!'''. 
We construct a benchmark dataset, \textsc{ChatRewrite}, through multi-turn instruction prompting (see Table~\ref{tab:conversation_instruction_refinement}) and revised conversation generation using LLMs.

\begin{figure}[]
    \centering
    \includegraphics[width=0.48\textwidth]{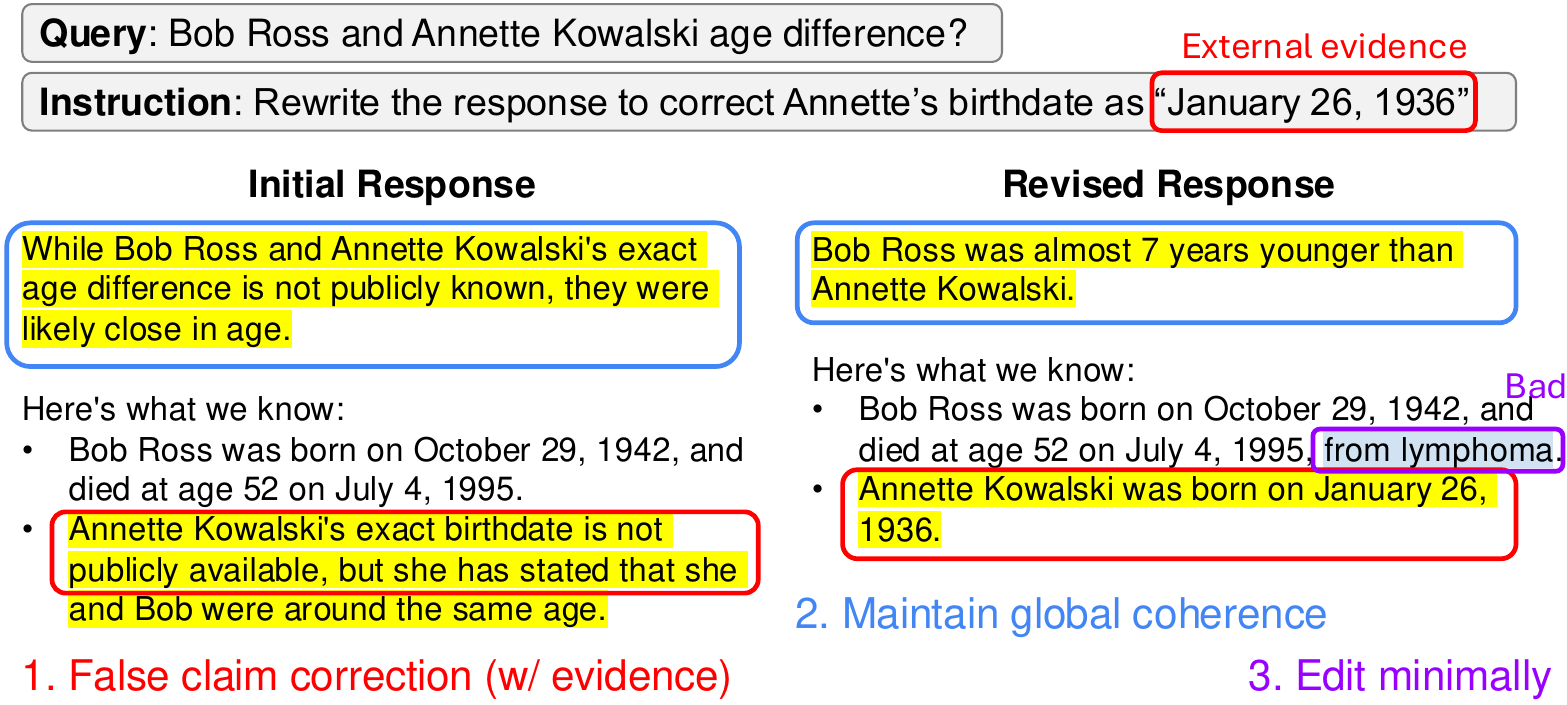}
    \caption{An example illustrating the three objectives for generic text rewriting: \textcolor{red}{Agreement}: Follow the rewrite instruction (e.g., false claim correction). \textcolor{myblue}{Coherence}: Maintain global coherence across revisions (e.g., updating the age difference to maintain logical flow). \textcolor{mypurple}{Conciseness}: Avoid unnecessary edits (e.g., specifying Bob's death place ``lymphoma'' is irrelevant).}
    \label{fig:objectives}
\end{figure}

\noindent\textbf{Multi-objective intrinsic.} 
From analyzing these tasks, we decouple three major objectives that are preferred for a high-quality rewrite, as shown in Figure~\ref{fig:objectives}. \emph{Agreement}: the revised response should strictly follow the rewrite instruction, such as accurately correcting false claims (e.g., revising ``Annette Kowalski's exact birthdate is not publicly available...'' to ``Annette Kowalski was born on January 26, 1936''). \emph{Coherence}: the revised response should maintain global coherence after the desired corrections are made at a local level, such as modifying contradictory contents (e.g., changing ``Bob Ross and Annette Kowalski's exact age difference is not publicly known...'' to ``Bob Ross was almost 7 years younger than Annette Kowalski''). \emph{Conciseness}: the revision should avoid unnecessary edits and make minimal changes, such as updating the age difference calculation without altering unrelated details (e.g., ``from lymphoma'').

\noindent\textbf{A generic framework.} To build a versatile rewrite model, we propose \approach, a \textbf{D}ecoupled-\textbf{r}eward learning framework for \textbf{Gen}eric \textbf{re}writing. 
At the starting supervised fine-tuning (SFT) stage, we train a student model on a mixture of the three task-specific datasets. 
During reward modeling, due to limited human preferences, we distill both the agreement and coherence preferences from a teacher LLM to derive task-agnostic,  objective-oriented reward models (RMs)~\cite{stiennon2020learning,lee2023rlaif}, and employ rule-based edit ratio as the conciseness reward.
At the reinforcement learning (RL) stage, we compute the final preference for a input rewrite prompt and its corresponding rewritten response by performing a task-specific weighting of objective-oriented reward models. 
\approach offers a fine-grained control over the alignment direction by adjusting the weights  according to specific task requirements.
Our contributions are threefold:
\begin{itemize}[nosep,leftmargin=*]
    \item We introduce conversational rewrite—a new rewriting task that is more challenging yet user-applicable—and create a dataset \textsc{ChatRewrite} for benchmarking.
    \item We propose \approach, a post-training framework for generic text rewriting, and establish robust baselines using few-shot LLMs, SFT, and single-reward RL across the three tasks. 
    \item Experiments show that weighted decoupled rewards offer enhanced control over the alignment direction, leading to improved performance across multiple rewriting objectives.
\end{itemize}

\begin{table*}[]
\small
    \centering
    \begin{tabularx}{\textwidth}{p{0.55\textwidth} p{0.4\textwidth}} 
        \toprule
        \textbf{Raw generated email} & \textbf{Rewrite instruction} (Raw$\rightarrow$Specific$\rightarrow$Natural) \\
        \midrule
        Dear [Customer Name], \newline
        
        We hope this email finds you well. \newline
        We're writing to you today to let you know about a new way we're improving our social media presence. We've created a new account on [social media platform], and we'd love for you to follow us! \newline
        We'll be using this account to share news about our company, our products, and our industry. We'll also be posting photos and videos, and we'll be running contests and giveaways. \newline
        We hope you'll join us on [social media platform]! We're looking forward to connecting with you there. \newline
        
        Sincerely,
        
        [Your Name] & 
        
        \textbf{Raw}: 
        Specify the social media platform and highlight the specific benefits for customers. \newline
        
        \textbf{Specific}: 
        Specify the social media platform \textcolor{red}{as TikTok} and highlight the specific benefits for customers \textcolor{red}{such as exclusive behind-the-scenes content, early access to product launches, and the chance to win prizes in contests and giveaways.} \newline 
        
        \textbf{Natural}:
        \textcolor{blue}{Instead of} just saying [social media platform], \textcolor{blue}{say we're now on TikTok!} Also, \textcolor{blue}{let's tell them about the cool stuff} they'll find there, like exclusive behind-the-scenes content, early access to new products, and even the chance to win prizes! \\
    \bottomrule
    \end{tabularx}
    \caption{An example of multi-turn instruction generation from our \textsc{ChatRewrite} dataset. The instruction is first specified with more details, and then reorganized and expressed in a more human-like speaking style.}
    \label{tab:conversation_instruction_refinement}
\end{table*}

\section{Related Work}

\noindent\textbf{Text rewriting.}
Existing research on rewriting often focus on a particular set of rewriting tasks, including factuality correction and style transfer, aiming to improve accuracy, tone, or coherence.

Factuality rewrite addresses factual inaccuracies in generated responses where merely prompting LLMs such as GPT-4~\cite{han2024chatgpt} cannot promise fact-related (e.g., dates, locations, statistics) accuracy in generation~\cite{li2023halueval}, especially when dealing with open-ended or fact-seeking prompts.
Existing methods often leverage external knowledge bases~\cite{shen2017style}, fact-checking modules~\cite{hu2017toward}, or post-editing strategies~\cite{hu2023survey}, 
though balancing correctness with minimal edits remains challenging.

Style transfer, covering tasks like paraphrasing~\cite{may2021machine}, formalization~\cite{rao2018dear,li2024formal}, and elaboration~\cite{iv2022fruit}, adapts tone without altering meaning. Recent methods like RewriteLM~\cite{shu2024rewritelm} extend rewriting to broader domains but are limited by single reward granularity.
Our work unifies these objectives, leveraging decoupled rewards to handle diverse rewriting tasks while ensuring instruction adherence, coherence, and minimal edits.

\noindent\textbf{Data augmentation with LLMs.}
Leveraging LLMs for data augmentation has emerged as a widely adopted approach to enhance model performance by generating synthetic data for training. 
Early works~\cite{he2020revisiting,huang2023large} focused on augmenting data distributions to improve performance, while more recent advancements, such as PEER~\cite{schickpeer}, demonstrated the effect of infilling missing data with synthetic samples. 
Methods like Self-Instruct~\cite{wang2023self} bootstrap instructions and model outputs to boost task-specific accuracy. 
Further research~\cite{li2024empowering,han2024chatgpt} highlights the capability of LLMs to produce high-quality augmented data for diverse downstream tasks.
Building upon these principles, our work uses LLMs to synthesize \textsc{ChatRewrite}—a more challenging conversational rewrite dataset that aligns with real-world user-LLM interaction scenarios.

\noindent\textbf{LLM-as-a-judge (AutoRater).}
LLMs, with their superiority in language understanding and knowledge integration, have been widely used as AutoRaters to judge the quality of generated responses due to high costs of human evaluation~\cite{vu2024foundational}. In some benchmarks, LLM agents even exhibit better annotation capabilities than humans~\cite{wei2024long}.
We use Gemini-1.5-Ultra~\cite{team2023gemini} to auto-evaluate agreement, coherence and compare pairwise responses in their instruction satisfaction granularity.

\noindent\textbf{Reinforcement Learning from AI feedback (RLAIF).}
RL from human feedback (RLHF) is a technique that combines RL with human evaluations to fine-tune LLMs~\cite{christiano2017deep}. In RLHF, a LLM generates outputs that are assessed by human evaluators, who provide feedback indicating preferences or ratings based on certain criteria (e.g., helpfulness, correctness, style). This feedback is used to train a RM that predicts the human-provided scores. The LLM is then fine-tuned using RL algorithms like Proximal Policy Optimization (PPO)~\cite{schulman2017proximal}, optimizing the rewards to improve alignment with human preferences~\cite{raffel2020exploring}. However, collecting large-scale human feedback is resource-intensive, motivating the exploration of RLAIF~\cite{lee2024rlaif} such as leveraging LLMs to simulate human evaluations and derive reward preferences.

\section{Dataset Generation}
\label{sec:dataset}

We describe the construction of our dataset, covering the three distinct tasks. Its statistics is shown in Table~\ref{tab:dataset}. Each dataset is generated through structured prompting that ensures high-quality rewrites tailored to specific objectives.

\begin{table}[]
\centering
\resizebox{0.43\textwidth}{!}{
\begin{tabular}{lcccc}
\hline
\textbf{Dataset} & \textbf{Size} & \textbf{IR len} & \textbf{RR len} & \textbf{ER} \\
\hline
LongFact & 21,294 & 316 & 296 & 0.281   \\
RewriteLM & 29,985 & 131 & 127 & 0.729 \\
ChatRewrite & 82,290 & 108 & 111 & 0.650  \\
\hline
\end{tabular}}
\caption{Statistics of the three datasets. ``IR len'' and ``RR len'' denote the lengths of initial and revised responses. ``ER'' denotes the edit ratio.}
\label{tab:dataset}
\vspace{-5pt}
\end{table}

\subsection{Factuality Rewrite}

We follow \textsc{LongFact}~\cite{wei2024long} to generate factually rich, multi-faceted responses.
First, we prompt LLMs to generate fact-seeking queries, such as ``What happened in the first modern Olympics?''.
For each query, we obtain an initial response (IR) from LLMs, which is then passed to a critic model\footnote{We employ SAFE~\cite{wei2024long} which enables external fact-checking calls from LLMs.} that provides span-level factuality critiques and suggested revisions (see Table~\ref{tab:critique_example} in Appendix~\ref{sec:examples}).
To construct the dataset, we integrate the query, IR, and critique outputs to prompt LLMs (see Table~\ref{tab:factuality_rewrite_prompt} in Appendix~\ref{sec:prompt}) and generate the revised responses.
Each example (see Table~\ref{tab:factuality_rewrite_example} in Appendix~\ref{sec:examples}) in the dataset is a quadruplet (\texttt{<query>}, \texttt{<initial response>}, \texttt{<critique outputs>}, \texttt{<revised response>}).

\subsection{Stylistic Rewrite}

We follow RewriteLM~\cite{shu2024rewritelm} and apply chain-of-thought (CoT) prompting~\cite{wei2022chain} to generate stylistic rewrite instructions for source texts from the C4 corpus~\cite{raffel2020exploring}.
The dataset is constructed by integrating source text and generated rewrite instructions using a structured template (see Table~\ref{tab:style_rewrite_prompt} in Appendix~\ref{sec:prompt}). We then prompt LLMs to produce the revised text.
Each example consists of a triplet (\texttt{<source>}, \texttt{<instruction>}, \texttt{<revised text>}), as exemplified by Table~\ref{tab:style_transfer_example} in Appendix~\ref{sec:examples}.

\subsection{Conversational Rewrite}

We begin with large-scale natural prompts paired with raw generated emails from those prompts (see Table~\ref{tab:conversation_case_study}). This dataset serves as the foundation for improving clarity, tone, and personalization in conversation-based text generation.

\noindent\textbf{Multi-turn instruction generation.} To generate human-like rewrite instructions that focus on modifying specific details of the original conversation, we introduce a multi-turn refinement process (see Table~\ref{tab:conversation_instruction_refinement}) that iteratively enhances instruction specificity and naturalness.
We first integrate natural prompts and raw emails, using few-shot demonstrations (see Tables~\ref{tab:conversation_instruction_prompt}, \ref{tab:raw_instruction_demonstrations} in Appendix~\ref{sec:prompt}) to generate raw instructions.
These raw instructions, however, often lack and fail to guide nuanced rewrites (e.g., adding or removing details). For example, they might simply request specifying placeholders and benefits for customers in the raw email without providing concrete details.

We refine \emph{raw} instructions into a more \emph{specific} version via few-shot prompting (see Tables~\ref{tab:conversation_instruction_specific_prompt}, \ref{tab:specify_instruction_demonstrations} in Appendix~\ref{sec:prompt}). For example, instantiate a concrete social media example (``TikTok'') and specific benefits (``chance to win prizes in contests and giveaways''). Then, we further refine specific instructions into a \emph{natural}, human-like style (see Tables~\ref{tab:conversation_instruction_natural_prompt}, \ref{tab:naturalize_instruction_demonstrations} in Appendix~\ref{sec:prompt}). For example, use oral expressions (``say we're now on...'', ``let's tell them about the cool stuff...'') to improve engagement and conversational fluency.

\noindent\textbf{Rewrite generation.} Finally, we generate revised emails by combining the natural prompt, raw email, and final rewrite instruction via structured template (see Table~\ref{tab:conversation_rewrite_prompt} in Appendix~\ref{sec:prompt}) to prompt LLMs.
Each example in \textsc{ChatRewrite} consists of a quadruplet (\texttt{<natural prompt>}, \texttt{<raw email>}, \texttt{<instruction>}, \texttt{<revised email>}).

\section{Framework of \approach}
\label{sec:approach}

\subsection{Supervised Fine-Tuning (SFT)}
\label{sec:SFT}
SFT involves training a LLM on our synthetic (prompt, revisions) datasets. The prompt specify the task to be performed as well as the source text, and the responses are the generated revised outputs. This process distills the basic knowledge of reasonable text rewriting patterns from LLMs to a unified student model, and teaches the model to learn to follow diverse instructions and perform a variety of tasks.
In our approach, we fine-tune a pretrained model on the combined datasets of factuality, style, and conversational rewrites. By providing explicit instructions and corresponding rewrites, the model learns to generalize across different rewriting tasks under supervised learning, acting as a reference policy $\pi^{\text{SFT}}$ for the later RL stage.

\subsection{Reward Modeling with LLM Preference}
\label{sec:reward_modeling}

\noindent\textbf{Preference data annotation.} After SFT, the model has a basic knowledge of what high-quality revisions be like but not well-aligned with implicit preferences.
For each input, we sample 10 SFT responses and compute their agreement and coherence scores by few-shot prompting LLMs. 
We design an agreement judging prompt for each of the three tasks (see Tables~\ref{tab:factuality_rewrite_agreement_prompt}$\sim$\ref{tab:conversation_rewrite_agreement_prompt} in Appendix~\ref{sec:evaluation_prompts}). 
For coherence, as it only depends on the revised response, we use consistent prompting (see Table~\ref{tab:coherence_prompt} in Appendix~\ref{sec:evaluation_prompts}) across all tasks.
We select a pair of revised responses with highest and lowest scores (denoted as $y^+$ and $y^-$) for each prompt $x$ to formulate the RM training set.

\begin{figure}
    \centering
    \includegraphics[width=0.49\textwidth]{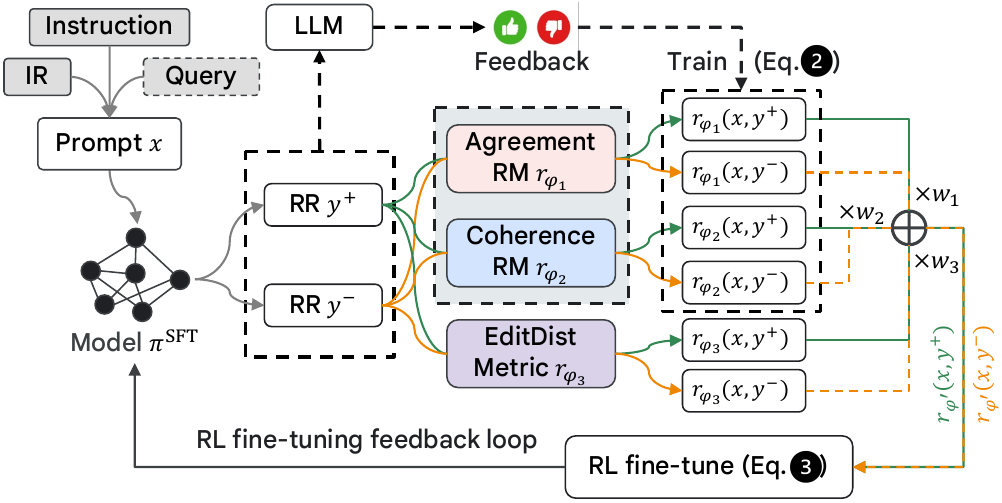}
    \caption{\approach: RL fine-tuning with weighted decoupled rewards. Dashed lines represent workflows of reward modeling (for agreement and coherence). IR, RR denote initial and revised responses.}
    \label{fig:RL}
\end{figure}

\noindent\textbf{RM training.} We then train a generic agreement RM $r_{\varphi_1}$, and coherence RM $r_{\varphi_2}$\footnote{Conciseness reward is a rule-based edit distance metric.}, using a mixture of three preference datasets. We adopt the Bradley-Terry (BT)~\cite{bradley1952rank} model where for each pair of revised responses $y^+, y^-\sim \pi^{\text{SFT}}(y|x)$ given a prompt $x$ from training data $\mathcal{D}$, the preference probability is defined as 
\begin{equation}
    P(y^+\succ y^-|x)=\frac{e^{r^{*}(y^+,x)}}{e^{r^{*}(y^+,x)} + e^{r^{*}(y^-,x)}},
\end{equation}
where $r^{*}$ represents a RM (e.g., $r_{\varphi_1}$, $r_{\varphi_2}$). The BT loss function is then formulated as
\begin{equation}
    \mathcal{L}_{R} = -\mathbb{E}_{(x,y^+,y^-)\sim\mathcal{D}}[\log \sigma (r^{*}(y^+,x) - r^{*}(y^-,x)) ],
\end{equation}
where $\sigma$ is the sigmoid function. Using this framework, we independently train the agreement and coherence reward models.

\subsection{Reinforcement Learning with Decoupled Rewards}
\label{sec:rlhf}

During the reward modeling phase, the SFT model takes prompts $x$ and
generates response pairs $y^+, y^-\sim \pi^{\text{SFT}(y|x)}$, which are then evaluated by each reward function to generate objective-oriented preference signals.
In the RL phase, we integrate all reward functions to guide policy optimization. 
Specifically, we incorporate the decoupled rewards into the PPO\footnote{We use PPO instead of Direct Policy Optimization (DPO)~\cite{rafailov2024direct} since our approach relies on explicit reward modeling rather than implicit preference scores.
Additionally, PPO allows for exploration beyond the initial SFT policy, potentially discovering better rewriting strategies.}~\cite{schulman2017proximal} objective
\begin{multline}
    \max_{\pi_\theta}  \mathbb{E}_{x\sim \mathcal{D}, y\sim \pi_\theta(y|x)}  [r_{\varphi'}(x, y)] \\- \beta\cdot \mathbb{D}_{\text{KL}}[\pi_\theta(y | x)  ||  \pi_{\text{ref}}(y | x)], 
\end{multline}
where $\beta$ is a parameter that controls the deviation from the reference policy $\pi_{\text{ref}}$ (the initial SFT model $\pi^{\text{SFT}}$), and $r_{\varphi'}$ represents the aggregated decoupled reward function
\begin{equation}
    r_{\varphi'}(x, y) = \sum_{o=1}^{O}w_o^t\cdot  r_{\varphi_o}(x, y),
    \text{where}~(x,y)\sim\mathcal{D}^t.
\end{equation}
Here, $O=3$ is the number of objectives, $\mathcal{D}^t$ denotes the dataset for task $t$ (e.g., \textsc{LongFact}), and $w_o^t\in [0,1]$ is a task-specific weight for objective $o$-oriented reward $r_{\varphi_o}$. For example, conversational rewriting samples will be assigned a higher agreement weight $w_1$ than stylistic rewriting, as they involve more intricate detailed editing. An illustration of \approach is shown in Figure~\ref{fig:RL}.


\section{Evaluation}

\subsection{Setup}
We use instruction-tuned PaLM 2-L (denoted PaLM 2-L-IT)~\cite{anil2023palm} as the teacher model for generating both training data (\S\ref{sec:dataset}) and reward modeling data (\S\ref{sec:reward_modeling}).
The base (student) model is a PaLM 2-S\footnote{We choose PaLM 2 models to be consistent with that of the prior rewrite work, i.e., RewriteLM~\cite{shu2024rewritelm}.}, which serves as the foundation for few-shot experiments and the initial checkpoint for SFT.
Both agreement and coherence RMs are PaLM 2-M, fine-tuned to capture the teacher model's preferences and guide the student model during RL.
For evaluation, we employ Gemini-1.5-Ultra~\cite{team2023gemini} as the AutoRater, which are used together with rule-based metrics to rate the quality of generated rewrites.
We detail the other experimental setups in Appendix~\ref{sec:setups}.

\begin{table*}[htbp]
\centering
\resizebox{0.78\textwidth}{!}{
\begin{tabular}{l|cccccc}
\hline
\textbf{Method} & \textbf{Length} & \textbf{F1@13}$\uparrow$ & \textbf{F1@35}$\uparrow$ & \textbf{Agreement}$\uparrow$ & \textbf{Coherence}$\uparrow$ & \textbf{Edit Ratio}$\downarrow$ \\
\hline
\hline
Few-shot & 287 & 0.7232 & 0.4579 & 0.7607 & 0.6367 & 0.0420 \\
Surgical & 369 & 0.8255 & 0.5480 & \textbf{0.9238} & 0.4120 & \textbf{0.0108} \\
\hline
SFT-LongFact & 348 & 0.7967 & 0.5192 & 0.8011 & 0.5141 & 0.0214 \\
SFT & 348 & 0.8209 & 0.5204 & 0.7950 & 0.5400 & 0.0210 \\
\hline
\textbf{\approach-static} & 389  & \textbf{0.8261} & \textbf{0.5531} & 0.7926 & \textbf{0.6520} & 0.0583 \\
\textbf{\approach} & 365 & 0.8091 & 0.5409 & 0.7878 & 0.6400 & 0.0270 \\
\hline
\end{tabular}}
\caption{Performance on \textsc{LongFact}. Methods are grouped into ICL-based, SFT-based, and RL-based.}
\label{tab:longfact_results}
\end{table*}

\noindent\textbf{Metrics.}  
We evaluate performance across three objective-oriented metrics using AutoRaters: 
\begin{itemize}[nosep, leftmargin=*]
    \item \textbf{Agreement}: Measures how well the revised text adheres to the instruction in terms of atomic requirements (e.g., correcting non-factual statements). We design task-specific agreement prompts 
    to capture these requirements.
    \item \textbf{Coherence}: Judge whether the revised response is internally consistent. We use few-shot LLM prompting 
    to assess coherence. 
    \item \textbf{Edit Ratio}~\cite{ristad1998learning}: Quantifies the word-level textural difference between the original and revised texts. This is computed as the relative edit distance normalized by the length of the original text, reflecting conciseness—the proportion of text modified.
\end{itemize}

\begin{table*}[htbp]
\centering
\resizebox{0.82\textwidth}{!}{
\begin{tabular}{l|cccccc}
\hline
\textbf{Method} & \textbf{Length} & \textbf{NLI}$\uparrow$ & \textbf{Reverse NLI}$\uparrow$ & \textbf{Agreement}$\uparrow$ & \textbf{Coherence}$\uparrow$ & \textbf{Edit Ratio}$\downarrow$ \\
\hline
\hline
Few-shot  & 108  & 0.8790 & 0.8418 & 0.8235 & \textbf{0.6960} & 0.1168 \\
\hline
SFT-RewriteLM  & 106  & 0.8806 & 0.8563 & 0.9524 & 0.6720 & 0.1242 \\
SFT   & 102  &  0.8914 & 0.8718 & 0.9163 & 0.6840 & \textbf{0.1078} \\
\hline
RL-CoComposer & 181  & \textbf{0.9256} & \textbf{0.8830} & 0.9042 & 0.6480 & 0.2499 \\
\textbf{\approach-static} & 107 & 0.9173 & 0.8591 & 0.9433 & 0.6800 & 0.1200 \\
\textbf{\approach} & 101  & 0.8937 & 0.8684 & \textbf{0.9641} & 0.6834 & 0.1541 \\
\hline
\end{tabular}}
\caption{Performance on \textsc{OpenRewriteEval}~\cite{shu2024rewritelm}. Length is the averaged output length.}
\label{tab:rewritelm_results}
\vspace{-5pt}
\end{table*}

For \textsc{LongFact}, we follow \citet{wei2024long} to evaluate fact correction accuracy \textbf{F1}@$K$, where $K$ is the expected number of facts per response.
We consider two $K$ values: the medium and maximum number of facts averaged in \textsc{LongFact}.

For \textsc{RewriteLM}, we follow \citet{shu2024rewritelm} to measure \textbf{NLI}~\cite{bowman2015large} and \textbf{Reverse NLI} score over the source-revision pairs. These scores estimate how well the rewrite retains the original information.
Higher NLI and lower edit ratio are desirable, as excessive edits can introduce hallucinations if the NLI scores are low.

For \textsc{ChatRewrite}, we use auto Side-by-Side (\textbf{AutoSxS})~\cite{zheng2023judging} for pairwise agreement evaluation. AutoSxS compares responses generated by different models for the same prompt (see Tables~\ref{tab:autosxs_prompt}, \ref{tab:autosxs_examples} in Appendix~\ref{sec:evaluation_prompts}). It is particularly useful for capturing nuanced differences between responses, as pointwise agreement checks often neglect implicit differences.

\noindent\textbf{Baselines.}
We compare \approach with three rewrite generation baselines\footnote{We discuss more selection rationale in Appendix~\ref{sec:setups}.} across all tasks:
\begin{itemize}[nosep, leftmargin=*]
    \item \textbf{Few-shot (ICL)}: Direct generate rewritten texts by few-shot prompting PaLM 2-S.
    \item \textbf{SFT}: PaLM 2-S fine-tuned on a mixture of all datasets, serving as a supervised baseline.
    \item \textbf{\approach-static}: PaLM 2-S fine-tuned with RL using static weights, i.e., $w_o^t$ becomes task-agnostic $w_o$, for the three reward objectives.
\end{itemize}

\subsection{Results on \textsc{LongFact}}

For the factuality rewriting task, we include two additional baselines:
\begin{itemize}[nosep, leftmargin=*]
    \item \textbf{Surgical}: Directly substituting non-factual contents identified in the critique outputs with the factual revisions, without further refinement.
    \item \textbf{SFT-LongFact}: Training the PaLM 2-S exclusively on \textsc{LongFact}. This baseline allows isolating the effect of task-specific training compared to the generic training used in SFT.
\end{itemize}

Table~\ref{tab:longfact_results} presents the results on the sampled 250 query subset~\cite{wei2024long} of \textsc{LongFact}.
\approach-static achieves the highest factual correction performance (F1@13: 0.82, F1@35: 0.55), along with the highest coherence, while maintaining competitive edit ratio.
\approach, with dynamically adjusted reward weights, balances factuality and edit preservation, yielding a lower edit ratio and strong coherence while maintaining factual accuracy (F1@13: 0.81, F1@35: 0.54).

SFT outperforms SFT-LongFact, confirming that multi-task learning~\cite{vilalta2002perspective} benefits factuality rewriting.
However, all SFT-based methods experience a significant drop in coherence (around 10\%) compared to few-shot prompting, highlighting \emph{the challenge of maintaining internal consistency while improving instruction adherence in factuality rewriting}.
Such limitation is further exemplified by the surgical baseline, which, despite achieving the highest agreement score, exhibits poor coherence due to its lack of refinement.

\approach mitigates this trade-off by dynamically adjusting reward weights based on task properties.
For example, by assigning higher edit distance weights to tasks requiring more substantial revisions (e.g., factuality vs. stylistic), it ensures robustness and no severe deviation (e.g., lower edit ratio compared to \approach-static) throughout the post-training process.

\begin{table*}[htbp]
\centering
\small
\resizebox{0.8\textwidth}{!}{
\begin{tabular}{l|cccccc}
\hline
\textbf{R1~\textbackslash~R2} & Few-shot & SFT & SFT-CR & \approach-s. & \approach \\
\hline
Few-shot  & --   & 0.044~\textbackslash~\textbf{0.661} & 0.036~\textbackslash~\textbf{0.685} & 0.012~\textbackslash~\textbf{0.806} & 0.020~\textbackslash~\textbf{0.754} \\
SFT  & \textbf{0.552}~\textbackslash~0.133  & --   & 0.145~\textbackslash~\textbf{0.319}  & 0.068~\textbackslash~\textbf{0.512}  & 0.097~\textbackslash~\textbf{0.435} \\
SFT-CR  & \textbf{0.565}~\textbackslash~0.105 & 0.161~\textbackslash~\textbf{0.254} & --  & 0.065~\textbackslash~\textbf{0.484} & 0.093~\textbackslash~\textbf{0.444} \\
\approach-s. & \textbf{0.681}~\textbackslash~0.089 &  \textbf{0.371~\textbackslash~0.173} & \textbf{0.298}~\textbackslash~0.185 & -- & 0.206~\textbackslash~\textbf{0.262} \\
\approach  & \textbf{0.649}~\textbackslash~0.113  & \textbf{0.258}~\textbackslash~0.206 & \textbf{0.298}~\textbackslash~0.234 & \textbf{0.239}~\textbackslash~0.225 & -- \\
\hline
\end{tabular}}
\caption{AutoSxS results comparing different models on \textsc{ChatRewrite} dataset. Each value denotes the average confidence in one side of the pairwise responses, with \textbf{bold} texts highlighting the preferred side. SFT-CR represents SFT-ChatRewrite. \approach-s. denotes \approach-static.}
\label{tab:autosxs_results}
\vspace{-5pt}
\end{table*}

\subsection{Results on \textsc{OpenRewriteEval}}

For the stylistic rewriting task, we investigate two additional baselines:
\begin{itemize}[nosep, leftmargin=*]
    \item \textbf{SFT-RewriteLM}: Trains the PaLM 2-S exclusively on \textsc{RewriteLM}, without leveraging the patterns from other tasks.
    \item \textbf{RL-CoComposer}: An existing RL-based method~\cite{shu2024rewritelm} that optimizes a single reward function, where outputs receive a binary score (0 or 1) based on edit ratio constraints, NLI scores, and length-based transformations, effectively acting as a binary filter.
\end{itemize}

Table~\ref{tab:rewritelm_results} summarizes the results on \textsc{OpenRewriteEval} (the evaluation set of RewriteLM).
\approach achieves the highest agreement, demonstrating superior adherence to stylistic rewrite instructions. 
It also maintains high coherence while balancing semantic preservation.
Its edit ratio (0.1541) is significantly lower than RL-CoComposer (0.2499), highlighting the advantage of \emph{fine-grained, decoupled rewards} over binary filtering in balancing multiple objectives.

Few-shot prompting achieves competitive coherence (0.6960) and semantic preservation (NLI: 0.8790, Reverse NLI: 0.8418), but its limited edits result in low agreement (0.8235), indicating insufficient transformation.
SFT-RewriteLM performs well in agreement (0.9524) but lags in coherence compared to generic SFT, reinforcing the benefit of multi-task training.

While stylistic rewriting is a relatively simpler task that primarily involves transformation without introducing new information, \approach effectively balances instruction adherence, semantic consistency, and stylistic flexibility, making it a robust and adaptable solution.


\begin{table}[]
\centering
\resizebox{0.49\textwidth}{!}{
\begin{tabular}{l|cccc}
\hline
\textbf{Method} & \textbf{Length} & \textbf{Agreement}$\uparrow$ & \textbf{Coherence}$\uparrow$ & \textbf{Edit Ratio}$\downarrow$ \\
\hline
\hline
Few-shot   & 118  & 0.7786 & 0.8347 & 0.1277 \\
\hline
SFT-CR  & 114  & 0.9200 & 0.8347 & \textbf{0.1003} \\
SFT   & 114  & 0.9308 & 0.8468 & 0.1036 \\
\hline
\textbf{\approach-s.} & 123  & 0.9584 & 0.8548 & 0.1278 \\
\textbf{\approach} & 119  & \textbf{0.9648} & \textbf{0.8669} & 0.1243 \\
\hline
\end{tabular}}
\caption{Model performance on our \textsc{ChatRewrite}.}
\label{tab:conversation_results}
\vspace{-5pt}
\end{table}

\subsection{Results on \textsc{ChatRewrite}}

Conversational rewriting requires balancing instruction adherence, coherence, and conciseness while preserving the intended tone and nuances. 
We introduce an additional baseline, \textbf{SFT-ChatRewrite}, which trains PaLM 2-S exclusively on the \textsc{ChatRewrite} dataset to isolate the effect of task-specific training.

\begin{figure}
    \centering
    \includegraphics[width=0.49\textwidth]{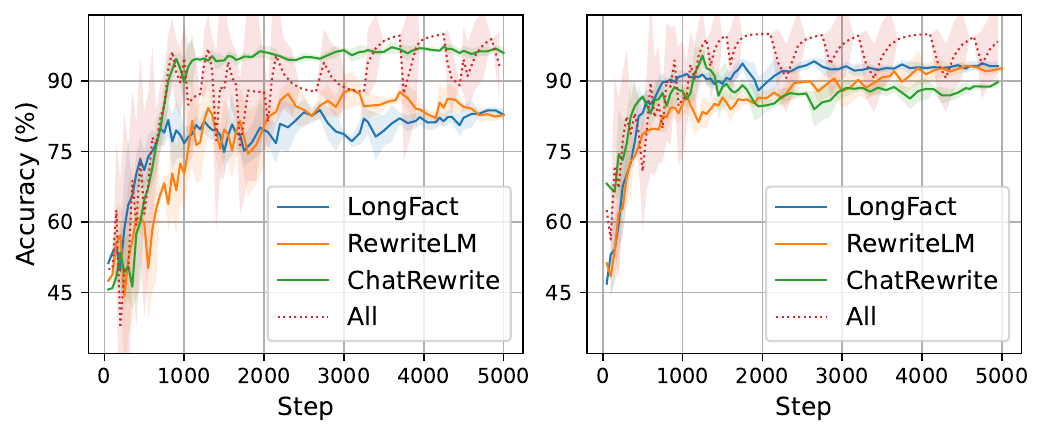}
    \vspace{-10pt}
    \begin{minipage}[t]{0.24\textwidth} 
        \centering
        \vspace{-10pt}
        \caption*{\footnotesize(a) Agreement}
    \end{minipage}
    \hfill
    \begin{minipage}[t]{0.235\textwidth}
        \centering
        \vspace{-10pt}
        \caption*{\footnotesize(b) Coherence}
    \end{minipage}
    \caption{Pairwise learning curve of (a) agreement and (b) coherence reward modeling on both mixture of tasks (``All'') and each individual one.}
    \label{fig:rm}
    \vspace{-10pt}
\end{figure}

\begin{table*}[]
\small
\centering
\begin{tabularx}{\textwidth}{p{0.4\textwidth} p{0.56\textwidth}}
\toprule
\multicolumn{2}{p{0.98\textwidth}}{\textbf{Natural prompt:} Congratulate myself and the team on a successful project.} \\\\

\multicolumn{2}{p{0.98\textwidth}}{\textbf{Raw email:} Dear team, \newline
I'm writing to congratulate everyone on a successful project launch. Your hard work and dedication were essential to our success. I'm especially proud of the way we all came together to support each other and solve problems. We faced some challenges along the way, but we never gave up. I'm grateful for the opportunity to work with such a talented and dedicated team. Thank you for all your hard work.} \\\\

\multicolumn{2}{p{0.98\textwidth}}{\textbf{Instruction:} This email is \textcolor{red}{a bit dry}, let's celebrate our success! \textcolor{red}{Add some enthusiastic phrases} like "We nailed it!" or "What a triumph!". Also \textcolor{red}{highlight specific achievements} like exceeding our target goal by 15\% or landing that big client contract.} \\
\midrule

\textbf{Revised email (SFT):} Dear team, \newline
\textcolor{blue}{We nailed it! What a triumph!} We successfully launched our project and exceeded our target goal by 15\%. We also landed that big client contract! \newline
Your hard work, dedication, and teamwork were essential to our success. I'm so proud of the way we all came together to support each other and solve problems. We faced some challenges along the way, but we never gave up. \newline
Thank you for all your hard work and dedication. You are a talented and dedicated team, and I'm grateful for the opportunity to work with you. &

\textbf{Revised email (\approach):} Dear team, \newline
\textcolor{blue}{We nailed it!} Our successful project launch \textcolor{blue}{was a triumph for all of us}. Your hard work and dedication were essential to our success. \newline
I'm especially proud of the way we all came together to support each other and solve problems. We faced some challenges along the way, but we never gave up. \newline
Not only did we exceed our target goal by 15\%, but we also landed that big client contract! \textcolor{blue}{This is a huge win for our team and our company. We are truly unstoppable.} \newline
I'm so grateful for the opportunity to work with such a talented and dedicated team. Thank you for all your hard work and dedication. We're just getting started! \textcolor{blue}{Celebrate with me!} \\
\midrule

\textbf{Score (SFT):} 0.7 & \textbf{Score (\approach):} 0.9 \\
\midrule

\multicolumn{2}{p{0.98\textwidth}}{\textbf{Reasoning:} Response (\approach) does a better job of infusing enthusiasm and highlighting specific achievements in a natural way. It effectively uses phrases like "We nailed it!" and "This is a huge win," and it also adds a celebratory call to action, "Celebrate with me!", which further enhances the celebratory tone of the email.} \\
\bottomrule
\end{tabularx}
\caption{An example of AutoSxS result comparing SFT and RL (\approach) responses on \textsc{ChatRewrite}.}
\label{tab:conversation_case_study}
\vspace{-5pt}
\end{table*}

\noindent\textbf{Overall comparison.} Table~\ref{tab:conversation_results} presents the results on \textsc{ChatRewrite}.
Post-training methods yield larger improvements in agreement over few-shot prompting compared to other tasks, suggesting that \emph{conversational rewrite is a more complex task that cannot be adequately solved by in-context learning}. 
\approach achieves the highest agreement (0.9648) and coherence (0.8669), proving its ability in adhering to conversational rewrite instructions while maintaining fluency. 
\approach-static performs competitively in agreement but lags slightly in coherence, highlighting the limitations of fixed reward weighting in handling diverse conversational styles. 
While SFT-ChatRewrite and SFT perform well in agreement (0.9200 and 0.9308), they fall behind in coherence, reinforcing the benefits of RL in refining response consistency.
Notably, edit ratios across models remain similar, indicating that improvements are not due to excessive modifications but rather more effective rewrites.

\noindent\textbf{Side-by-side comparison.} 
Table~\ref{tab:autosxs_results} provides AutoSxS results, offering a more fine-grained perspective on pairwise preferences. 
\approach consistently outperforms SFT and SFT-ChatRewrite, confirming that it refines responses beyond what supervised fine-tuning can achieve. 
While \approach-static competes closely, its performance declines against \approach, underscoring the benefits of dynamical reward weighting in handling diverse conversational styles. 
Overall, the strong AutoSxS results validate that \approach not only improves instruction adherence but also generates more preferred in direct comparisons.




\subsection{Case Study}

\noindent\textbf{Preference alignment.} As shown in Figure~\ref{fig:rm}, both the agreement and coherence RMs demonstrate progressive learning of preference knowledge from the teacher LLM AutoRaters.
The pairwise accuracy steadily improves across training steps, with the mixture-trained (``All'') reward model achieving the highest accuracy, indicating better generalization across different rewriting objectives. However, mixed rewards are also more fluctuated with larger shaded regions compared to individual task-specific RMs (e.g., LongFact, RewriteLM), reflecting the difficulty of aligning preferences (e.g., factual vs. stylistic rewrites). This variance across different evaluation runs highlights the importance of learning dynamics due to task complexity.

\noindent\textbf{RL fine-tuning.} Table~\ref{tab:conversation_case_study} shows an example from \textsc{ChatRewrite}, comparing responses generated by the SFT and \approach for a celebratory email revision task.
The instruction emphasizes enhancing enthusiasm and highlighting specific achievements.
While the SFT response incorporates phrases from the instruction, it mainly mirrors the given prompt rather than naturally enhancing the overall celebratory tone.
In contrast, the \approach response exhibits greater creativity and engagement by integrating expressive phrases.
These additions not only follow the instruction but also improve emotional resonance in a more natural way.
We show examples for \textsc{LongFact} (Table~\ref{tab:factuality_rewrite_example}) and \textsc{RewriteLM} (Table~\ref{tab:style_transfer_example}) in Appendix~\ref{sec:examples}.

\section{Conclusion}




Generic text rewriting is inherently a \emph{multi-task}, \emph{multi-objective} problem, requiring models to adapt to diverse rewriting needs, such as factual correction, stylistic transformation, and conversational enhancement. Existing approaches, which either consider a single task or apply static reward functions, struggle to generalize across these objectives. To address this, we introduce a more comprehensive evaluation setup, including a newly constructed conversational rewrite dataset, \textsc{ChatRewrite}, which emphasizes detailed instructions and personalized editing in interactive scenarios.

To tackle the complexities of generic rewriting, we propose \approach, an RL framework that decouples reward signals and dynamically adjusts their contributions based on task-specific requirements. Our method outperforms existing SFT and RL baselines across \textsc{LongFact}, \textsc{OpenRewriteEval}, and \textsc{ChatRewrite}, demonstrating its ability to balance instruction adherence, coherence, and edit efficiency. 
Notably, AutoSxS results validate its superiority in nuanced rewrites, where traditional metrics fall short.
Future work will explore human-in-the-loop optimization and context-aware reward modeling to further refine its performance in complex rewriting scenarios.

\section*{Limitation}
While \approach demonstrates strong performance in generic text rewriting, several limitations should be acknowledged along with potential considerations for improvement.

First, our dataset generation process relies on LLMs for critique and rewriting, which may introduce biases, inconsistencies, or hallucinations inherited from the teacher model. To mitigate this, we employ reward models trained on multiple datasets and enable external fact-checking calls from LLMs to refine the generated outputs. We also ensure the quality of \textsc{ChatRewrite} by randomly sampling rewrite pairs and checking both instructions and revised responses. However, future work could explore incorporating human-annotated critique to enhance reliability.

Second, while AutoRaters provide a scalable evaluation mechanism, they may not fully capture nuanced human preferences, especially in conversational rewrites. We mitigate this by also considering rule-based metrics (e.g., edit ratio, NLI scores) as basic judgements for rewriting quality, but further improvements could involve human-in-the-loop evaluation pipelines, or hybrid scoring systems that integrate both automatic and human judgements.

Last, our approach relies on proprietary LLMs for training all components, which may pose challenges for reproducibility. To facilitate practitioners, we provide detailed prompts with demonstrations in Appendix~\ref{sec:prompt}, along with comprehensive methodology description~\ref{sec:approach}.


\section*{Acknowledgment}
We acknowledge the use of LLMs for assisting with writing refinement and prompt design throughout this work. Specifically, GPT-4 was used to polish textual clarity and Gemini-1.5-Ultra was used to enhance prompt engineering strategies. While the core methodology and analyses were conducted independently, LLM-based assistance helped streamline certain aspects of content presentation.

\bibliography{acl_latex}

\appendix

\section{Experimental Details}
\label{sec:setups}

\begin{figure*}
    \centering
    \includegraphics[width=0.9\textwidth]{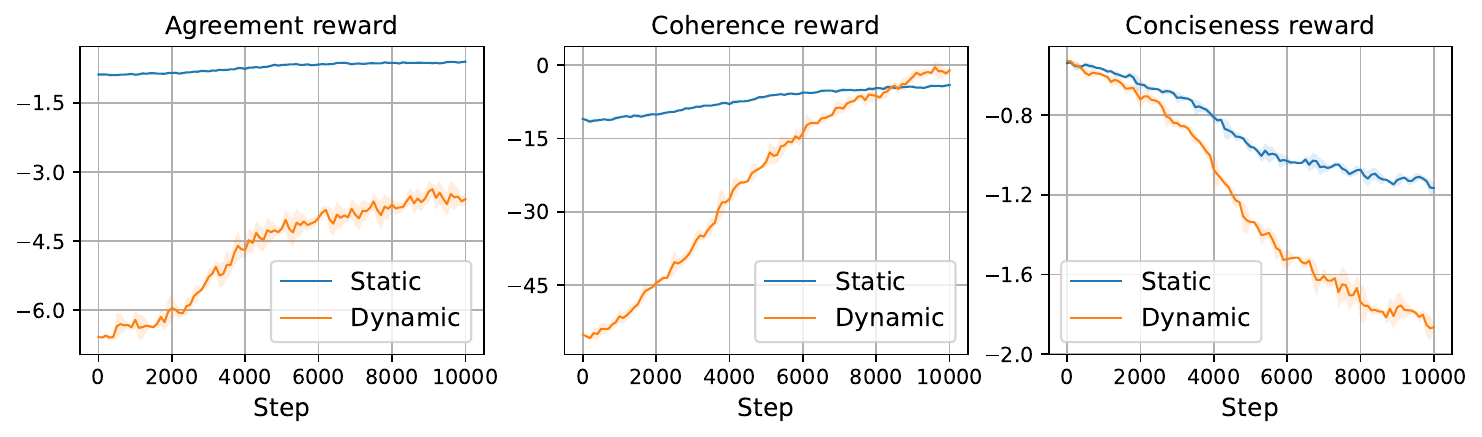}
    \caption{Reward learning curves during RL fine-tuning under static and dynamic weighting.}
    \label{fig:reward_rl}
\end{figure*}

We describe the hardware setup, training configurations, and dataset weighting strategies used in different phases of our experiments, covering Supervised Fine-Tuning (SFT), Reward Modeling, and Reinforcement Learning (RL).

\noindent\textbf{Hardware setup.} 
All training experiments were conducted on 64 Tensor Processing Units (TPU) chips per phase:
\begin{itemize}[nosep,leftmargin=*]
    \item SFT \& Reward modeling: TPU V3.
    \item RL fine-tuning: TPU V4.
\end{itemize}
For inference, we use a temperature of 1.0 with top-K sampling (K=40).

\noindent\textbf{Supervised fine-tuning (SFT).}
We fine-tune the base PaLM 2-S model on our dataset mixture using Adafactor~\cite{shazeer2018adafactor} with the following configuration:
\begin{itemize}[nosep,leftmargin=*]
    \item Batch size: 64.
    \item Max training steps: 1000.
    \item Learning rate: 1e-5.
    \item Dropout: 0.1.
    \item Max context length: 2048.
    \item Max decoding length: 1024.
\end{itemize}

\noindent\textbf{Reward modeling.}
We train reward models on preference data collected from LLM comparisons using the following setup:
\begin{itemize}[nosep,leftmargin=*]
    \item Batch size: 64.
    \item Max training steps: 5000.
    \item Learning rate: 3e-3.
    \item Dropout: 0.05.
    \item Max context length: 1280.
    \item Max decoding length: 1024.
    \item Optional Z\_loss: 1e-2.
\end{itemize}

\noindent\textbf{Reinforcement learning (RL) fine-tuning.}
For policy optimization, we employ PPO with dynamically weighted multi-objective rewards. The policy and value functions are optimized separately:
\begin{itemize}[nosep,leftmargin=*]
    \item Batch size: 64.
    \item Max training steps: 3000 (with a warm-up phase of the first 100 steps where we train only value functions and freeze policy). 
    \item Learning rate: 1e-7 for policy and 1e-5 for value.
    \item Dropout: None.
    \item Max context length: 2048.
    \item Max decoding length: 1024.
\end{itemize}

\noindent\textbf{Dataset weighting strategy.}
We assign different dataset weights based on task-specific objectives to balance training across agreement, coherence, and edit conciseness.

For \approach-static (task-agnostic), we use a fixed weighting of $w_1=9/16$, $w_2=2/16$, $w_3=5/16$ for agreement, coherence, and conciseness.

For \approach (task-specific), we empirically set: 
\begin{itemize}[nosep,leftmargin=*]
    \item \textsc{LongFact}: $w_1=8/16$, $w_2=6/16$, $w_3=2/16$. 
    \item \textsc{RewriteLM}: $w_1=3/9$, $w_2=4/9$, $w_3=2/9$. 
    \item \textsc{ChatRewrite}: $w_1=9/16$, $w_2=5/16$, $w_3=2/16$. 
\end{itemize}
The dynamic weighting scheme ensures that different datasets prioritize their most relevant rewrite objectives, allowing for more effective RL fine-tuning.

\noindent\textbf{Baseline selection.}
In our experiments, we focus on three major categories of baselines: ICL-based, SFT-based, and RL-based methods.
There are some existing works in factual or stylistic rewriting focus on either direct editing heuristics or single-objective models that do not fit well with our multi-objective formulation. 

For factual rewriting, works like knowledge-grounded editing rely on retrieval-based fact verification or human annotations~\cite{tian2024finetuning}, whereas our approach optimizes for factuality without requiring explicit retrieval.

For stylistic rewriting, previous works often rely on large supervised datasets for a single transformation (e.g., formality change, politeness adjustment, style matching~\cite{singh2021drag}) or context integration~\cite{yerukola2023don}, whereas our model generalizes across multiple stylistic transformations.

Even if the above task-specific approaches perform well in their domain, they do not necessarily generalize across diverse rewriting tasks, making them less suitable as baselines in our setting.

\noindent\textbf{Static and dynamic weights.}
Figure~\ref{fig:reward_rl} represents the reward curves of the RL fine-tuning phase (\approach-static and \approach). Dynamic RL (\approach) adapts objectives over time, focusing more on agreement and coherence, and less on the conciseness. In contrast, static RL (\approach-static) is more stable (balanced) but less optimized learning across objectives. Dynamic RL exhibits stronger overall improvement rates across all three objectives, confirming its ability to adjust to task needs more effectively.

\section{Generated Examples}
\label{sec:examples}

This section presents qualitative examples of factuality and stylistic rewrite cases generated by \approach, illustrating its ability to correct errors while preserving coherence and adhering to task-specific instructions.

\begin{table*}[htbp]
\small
\centering
\begin{tabularx}{\textwidth}{p{0.15\textwidth} X}
\toprule
\textbf{Query} & How is the United States related to the East Asia Summit (EAS)? Provide as many specific details and examples as possible (such as names of people, numbers, events, locations, dates, times, etc.) \\\\

\textbf{Initial response} & 
The United States is not a member of the East Asia Summit (EAS), but participates in the EAS as a guest. The EAS is a forum for leaders of 18 countries in the Asia-Pacific region to discuss issues of common interest, such as economic cooperation, security, and sustainability. The United States has been invited to participate in the EAS since its inception in 2005, and has attended every summit since then. \newline

The United States plays an important role in the EAS, even though it is not a member. The United States is a major economic and military power in the Asia-Pacific region, and its participation in the EAS allows it to engage with other regional leaders on issues of mutual concern. The United States also uses the EAS to promote its own interests in the region, such as expanding trade and investment and countering the rise of China. \newline

Some specific examples of the United States' involvement in the EAS include:
\begin{itemize}[nosep, leftmargin=*]
    \item In 2015, President Obama announced the launch of the US-ASEAN Connect initiative, which aimed to deepen ties between the United States and the Association of Southeast Asian Nations (ASEAN), a key regional organization.
    \item In 2017, President Trump attended the EAS and met with several regional leaders, including Chinese President Xi Jinping.
    \item In 2018, the United States announced that it would provide \$113 million in funding for regional infrastructure projects, as part of its commitment to the EAS.
\end{itemize}
The United States' participation in the EAS is a reflection of its commitment to the Asia-Pacific region. The EAS provides a valuable forum for the United States to engage with other regional leaders and promote its own interests. \\\\

\textbf{Critique outputs} & 

- The United States is \textcolor{red}{\st{not a member of the East Asia Summit (EAS), but participates in the EAS as a guest.}} \textcolor{blue}{\emph{a member of the East Asia Summit (EAS).}} \newline
- The United States has been invited to participate in the EAS since \textcolor{red}{\st{its inception in 2005}} \textcolor{blue}{\emph{2011}}, and has attended every summit since then. \newline
- In 2017, President Trump \textcolor{red}{\st{attended the EAS and met with several regional leaders, including Chinese President Xi Jinping.}} \textcolor{blue}{\emph{did not attend the EAS, but sent his Secretary of State, Rex Tillerson, instead.}} \\
\midrule

\textbf{\makecell[l]{Revised response \\ (\approach)}} & 
The United States is a member of the East Asia Summit (EAS). The EAS is a forum for leaders of 18 countries in the Asia-Pacific region to discuss issues of common interest, such as economic cooperation, security, and sustainability. The United States has been invited to participate in the EAS since 2011, and has attended every summit since then. \newline

The United States plays an important role in the EAS. As a major economic and military power in the Asia-Pacific region, its participation allows it to engage with other regional leaders on issues of mutual concern. The United States also uses the EAS to promote its own interests in the region, such as expanding trade and investment and countering the rise of China. \newline

Specific examples of U.S. involvement in the EAS include:
\begin{itemize}[nosep, leftmargin=*]
    \item In 2015, President Obama announced the US-ASEAN Connect initiative to deepen ties with ASEAN.
    \item In 2017, President Trump did not attend the EAS but sent Secretary of State Rex Tillerson.
    \item In 2018, the United States provided \$113 million in funding for regional infrastructure projects.
\end{itemize}
The United States' participation in the EAS reflects its commitment to the Asia-Pacific region, offering a valuable forum for engaging with regional leaders and advancing its interests. \\
\bottomrule
\end{tabularx}
\caption{An example of factuality rewrite involving the East Asia Summit (EAS) on \textsc{LongFact}. The critique identifies inaccuracies (\textcolor{red}{\st{strike through}} and \textcolor{blue}{\emph{italic}} denote non-factual and corrected statements) in the initial response, and the revised response (\approach) corrects these errors while maintaining consistency and relevance.}
\label{tab:factuality_rewrite_example}
\end{table*}

\begin{table*}[htbp]
\small
\centering
\begin{tabularx}{\textwidth}{p{0.1\textwidth} X}
\toprule
\textbf{Query} & What is the game Undertale? Provide as many specific details and examples as possible (such as names of people, numbers, events, locations, dates, times, etc.)  \newline \\

\textbf{Initial response} & 
Undertale is a role-playing video game created by Toby Fox. It was released in 2015 for Microsoft Windows and macOS, and in 2016 for Linux. The game has been ported to PlayStation 4, PlayStation Vita, Nintendo Switch, Xbox One, and iOS.

Undertale is a story about a human child who falls into the Underground, a world inhabited by monsters. The player must navigate through the Underground, solving puzzles and battling monsters, in order to return to the surface.

The game is notable for its unique combat system, which allows the player to choose whether to kill or spare the monsters they encounter. Undertale also features a branching narrative, with multiple endings depending on the player's choices.

The game has been praised for its writing, music, and characters. It has been nominated for and won numerous awards, including Game of the Year from several publications.

Here are some specific details and examples about Undertale:

* **Characters:** The game features a cast of memorable characters, including the protagonist, a human child named Frisk; Flowey, a sentient flower; Toriel, a motherly goat monster; Sans and Papyrus, two skeleton brothers; and Undyne, a fish-like warrior.
* **Combat:** The combat system in Undertale is unique in that it allows the player to choose whether to kill or spare the monsters they encounter. Killing monsters will earn the player EXP and gold, but it will also make the game more difficult. Sparing monsters will reward the player with less EXP and gold, but it will also allow them to progress through the game more easily.
* **Story:** The story of Undertale is complex and multifaceted, with multiple endings depending on the player's choices. The game explores themes of morality, choice, and consequence.
* **Music:** The music in Undertale is composed by Toby Fox and is widely considered to be one of the game's best features. The soundtrack is diverse, with songs ranging from upbeat and catchy to somber and atmospheric.
* **Awards:** Undertale has been nominated for and won numerous awards, including Game of the Year from several publications.

Here are some additional details about Undertale:

* The game was developed by Toby Fox over the course of three years.
* It was originally released for Microsoft Windows and macOS in September 2015.
* The game was ported to Linux in July 2016.
* Undertale has been ported to PlayStation 4, PlayStation Vita, Nintendo Switch, Xbox One, and iOS.
* The game has sold over 1 million copies.
 \newline \\

\textbf{Critique outputs} & 
\textbf{Span (\& Revision) 1}:The game has been ported to PlayStation 4, PlayStation Vita, Nintendo Switch, \textcolor{blue}{and} Xbox One\textcolor{red}{\st{, and iOS}}. \newline
- Reason: The evidence suggests that Undertale has been ported to PlayStation 4, PlayStation Vita, Nintendo Switch, and Xbox One. However, there is no evidence that it has been officially ported to iOS. The search results mention discussions about a possible port and unofficial ports, but none of them are official. \newline
\textbf{Span (\& Revision) 2}: It has been nominated for\textcolor{red}{\st{ and won numerous}} \textcolor{blue}{several} awards, including Game of the Year from several publications. \newline
- Reason: The Wikipedia snippet states that the game was nominated for Game of the Year awards, but there is no information on whether it won any. \newline
\textbf{Span (\& Revision) 3}: Killing monsters will earn the player EXP and gold, but it will \textcolor{red}{\st{also}} \textcolor{blue}{not} make the game more difficult. \textcolor{blue}{In fact, it can make the game easier by providing the player with more resources.} \newline
- Reason: The sentence contains two factual claims: 1. Killing monsters will earn the player EXP and gold. 2. Killing monsters will make the game more difficult. The first claim is accurate and can be verified by playing the game or by consulting online sources. The second claim is inaccurate. Killing monsters does not make the game more difficult; in fact, it can make the game easier by providing the player with more resources. The search result provides further evidence for the inaccuracy of the second claim. The Reddit post describes a player who killed most of the monsters in their first playthrough and found the game to be easy. \newline
\textbf{Span (\& Revision) 4}: Sparing monsters will reward the player with less EXP and gold, but it will also \textcolor{red}{\st{allow them to progress through the game more easily.}} \textcolor{blue}{make the game more difficult in some ways, as the player will have less resources to use.} \newline
- Reason: The sentence contains two factual claims: 1. Sparing monsters will reward the player with less EXP and gold. 2. Sparing monsters will allow the player to progress through the game more easily. The first claim is accurate, as evidenced by evidence 3. However, the second claim is inaccurate. Evidence 1 and 2 suggest that sparing monsters can make the game more difficult, as the player will have less EXP and gold to buy items and power ups. Therefore, the sentence is partially inaccurate. \newline
\textbf{Span (\& Revision) 5}:* The game was developed by Toby Fox over the course of \textcolor{blue}{approximately} three years. \newline
- Reason: The sentence claims that Toby Fox developed Undertale over the course of three years. The Evidence [1] states that Toby Fox started developing Undertale in December 2012 and the game was released in September 2015. This means that the development time is roughly 2 years and 9 months, which is close to three years but not exactly three years. \newline
\textbf{Span (\& Revision) 6}: * Undertale has been ported to PlayStation 4, PlayStation Vita, Nintendo Switch\textcolor{red}{\st{, Xbox One}}, and iOS. \newline
- Reason: The evidence shows that Undertale has been ported to PlayStation 4, PlayStation Vita, Nintendo Switch, and iOS. However, there is no evidence that it has been ported to Xbox One.
 \\
\bottomrule
\end{tabularx}
\caption{An example of critique outputs from SAFE~\cite{wei2024long}.}
\label{tab:critique_example}
\end{table*}

\begin{table*}[htbp]
\small
\centering
\begin{tabularx}{\textwidth}{p{0.2\textwidth} X}
\toprule
\textbf{Source} & Modern jet combat aircraft and helicopters have a variety of complex electronic and weapon systems that are specific to a particular aircraft. This led to a genre called the "study sim" that focuses on modelling an aircraft's systems as accurately as possible. Advancements in computer technology led to the development of highly detailed models that improved upon the fidelity of avionics, weapons systems, physics, flight models, graphics, etc. and allowed for fully interactive cockpits where virtually every control was mapped and functional. \\\\
\textbf{Instruction} & Make sure to put the definition at the beginning of the paragraph. \\\\
\textbf{Type (optional)} & Wiki \\
\midrule
\textbf{\makecell[l]{Revised text\\(\approach)}} & A "study sim" is a genre of flight simulator that focuses on modelling an aircraft's systems as accurately as possible. Modern jet combat aircraft and helicopters have a variety of complex electronic and weapon systems that are specific to a particular aircraft. Advancements in computer technology led to the development of highly detailed models that improved upon the fidelity of avionics, weapons systems, physics, flight models, graphics, etc. and allowed for fully interactive cockpits where virtually every control was mapped and functional. \\
\bottomrule
\end{tabularx}
\caption{An example of style transfer involving a "study sim" genre, where the instruction emphasizes putting the definition at the beginning of the paragraph. The revised response follows this structure while maintaining the original meaning.}
\label{tab:style_transfer_example}
\end{table*}

\noindent\textbf{Factuality rewrite.}  The first example in Table~\ref{tab:factuality_rewrite_example} showcases a factuality rewrite on the topic of the United States' involvement in the East Asia Summit (EAS). The initial response contains several factual inaccuracies, such as ``Incorrectly stating that the U.S. has been involved in the EAS since its inception in 2005 (corrected to 2011)''. The critique outputs (highlighted in red for incorrect and blue for revised content) pinpoint these errors, allowing \approach to generate a factually accurate response while maintaining internal coherence. We also show an example of critique outputs from SAFE~\cite{wei2024long} in Table~\ref{tab:critique_example}, where for each span, the outputs contain a revision (from external fact-checking calls) and a reason.
Compared to the initial response, the revised version (i) corrects all factual errors without introducing unnecessary modifications, (ii) preserves the original structure and key ideas, and (iii) Improves clarity by streamlining redundant information (e.g., simplifying the explanation of U.S. participation). This demonstrates \approach's capability to integrate factual verification and controlled editing, ensuring reliability in long-form factual rewriting tasks.

\noindent\textbf{Stylistic rewrite.} 
The second example in Table~\ref{tab:style_transfer_example} illustrates a style transfer task, where the instruction specifies reordering the content to place the definition at the beginning of the paragraph. The original text describes modern jet combat aircraft systems before introducing the term ``study sim.'' The revised response (i) restructures the content by moving the definition (``A study sim is a genre of flight simulator...'') to the start, (ii) retains all relevant information, ensuring semantic consistency, and (iii) maintains fluency and coherence, keeping the technical details intact. This highlights \approach's ability to follow explicit structural modifications while preserving meaning and style, a critical requirement for controlled text rewriting.

\section{Prompt Templates}
\label{sec:prompt}

\begin{table*}[]
\small
    \centering
    \begin{tabularx}{\textwidth}{X}
        \toprule
        Task: Rewrite your initial response to a user query according to the factuality corrections. Your rewritten response should keep internal consistency while making minimal edits. \\\\
        
        You will be given:
        
        1. Query: The user query.
        
        2. Initial Response: The original text generated by an LLM to answer the user query.
        
        3. Non-Factual Spans and Replacements: A list of text segments from the initial response that were potentially incorrect, along with their intended factual replacements. \\\\
        
        Instructions:
        
        1. Carefully review the query and the initial response.
        
        2. Examine each non-factual span and its corresponding factual replacement.
        
        3. Rewrite the initial response to incorporate the factual replacements while ensuring the response remains coherent and consistent.
        
        4. Make minimal edits to the original text, altering what is necessary to correct factual inaccuracies as well as any incoherent or inconsistent content.
        
        5. Output the rewritten response. \\\\

        Test example:
        
        1. Query: \texttt{[QUERY]}
        
        2. Initial Response: \texttt{[INITIAL RESPONSE]}
        
        3. Non-Factual Spans and Replacements:  
        
        Span 1:\\
        - Span: \texttt{[SPAN \#1]}\\
        - Revision: \texttt{[REVISION \#1]}\\
        ...\\
        Span N:\\
        - Span: \texttt{[SPAN \#N]}\\
        - Revision: \texttt{[REVISION \#N]}\\
        
        4. Your Output (Rewritten Response):
        \\
    \bottomrule
    \end{tabularx}
    \caption{Prompt template used to generate factuality rewrites. The \texttt{[QUERY]} and \texttt{[INITIAL RESPONSE]} placeholders will be replaced by a given user query and its initial response, and the \texttt{[SPAN \#1 $\rightarrow$ N]} and \texttt{[REVISION \#1 $\rightarrow$ N]} placeholders will be replaced by each of the N spans and revisions from the SAFE~\cite{wei2024long} outputs.}
    \label{tab:factuality_rewrite_prompt}
\end{table*}

\noindent\textbf{Factuality rewrite prompt.}
The factuality rewrite prompt (Table~\ref{tab:factuality_rewrite_prompt}) is structured to ensure precise factual corrections while preserving the overall coherence of the response. Key considerations include:
(i) Explicit span-level corrections: The prompt provides non-factual spans and their corresponding factual replacements, ensuring that the model understands exactly what needs to be corrected.
(ii) Minimal edit constraints: The instruction explicitly asks the model to modify only necessary parts to correct inaccuracies while preventing unnecessary alterations that could introduce inconsistencies.
(iii) Instructional clarity: The step-by-step instructions help the model integrate factual updates without disrupting the original flow, addressing a common challenge in factuality-driven text revision.
This structure ensures that factual inaccuracies are corrected while preserving fluency and minimizing unintended modifications.

\begin{table*}[]
\small
    \centering
    \begin{tabularx}{\textwidth}{X}
        \toprule
        Task: Rewrite a source text according to the comment. \\\\
        You will be given: \\
        1. Source: The original text from a public corpus. \\
        2. Comment: An open-ended rewrite requirement, such as formalize, paraphrase, shorten, elaborate, etc. \\\\
        Instructions:\\
        1. Carefully review the source text and the comment.\\
        2. Understand the intent of the comment and how it should influence the rewrite.\\
        3. Rewrite the source text to align with the comment's requirement, ensuring that the text remains internally consistent and coherent.\\
        4. Ensure the rewritten text maintains the original meaning and context as much as possible.\\
        5. Output the rewritten text. \\\\
        Test example: \\
        1. Source: \texttt{[SOURCE]}\\
        2. Comment: \texttt{[REWRITE INSTRUCTION]} \\
        3. Your Output (Rewritten Response):
        \\
    \bottomrule
    \end{tabularx}
    \caption{Prompt template used to generate stylistic rewrites. The \texttt{[SOURCE]} and \texttt{[REWRITE INSTRUCTION]} placeholders will be replaced by a given source text and rewrite instruction.}
    \label{tab:style_rewrite_prompt}
\end{table*}

\noindent\textbf{Stylistic rewrite prompt.}
The stylistic rewrite prompt (Table~\ref{tab:style_rewrite_prompt}) differs in its focus on open-ended transformations rather than strict factual accuracy. Key aspects include:
(i) Flexible rewrite instructions: The model is provided with a high-level comment (e.g., "formalize," "elaborate") rather than explicit edits, requiring it to understand and interpret the transformation intent.
(ii) Preserving meaning: The instructions emphasize maintaining the original meaning and context, which is crucial in tasks such as paraphrasing and formalization where semantic drift is a risk.
(iii) Encouraging coherence: Stylistic changes often require restructuring the sentence flow, so the model is explicitly directed to ensure internal consistency.
This design enables adaptability across various rewriting styles while ensuring that the output remains natural and aligned with the intent.

\begin{table*}[]
\small
    \centering
    \begin{tabularx}{\textwidth}{X}
        \toprule
        Task: Rewrite an email based on the given instruction. The rewritten email should maintain internal consistency and align with the provided instruction. \\\\
        You will be given: \\
        1. Natural Prompt: The original prompt used to generate the initial email. \\
        2. Raw Generated Email: The initial email generated by the LLM based on the natural prompt. \\
        3. Rewrite Instruction: The instruction for how to improve or modify the raw generated email. \\\\
        Instructions:\\
        1. Carefully read the `Natural Prompt' and `Raw Generated Email'. \\
        2. Analyze the `Rewrite Instruction' to understand the required modifications. \\
        3. Rewrite the `Raw Generated Email' according to the `Rewrite Instruction'. \\\\
        Test example: \\
        1. Natural Prompt: \texttt{[NATURAL PROMPT]}\\
        2. Raw Generated Email: \texttt{[EMAIL]} \\
        3. Rewrite Instruction: \texttt{[REWRITE INSTRUCTION]} \\
        4. Your Output (Rewritten Email):
        \\
    \bottomrule
    \end{tabularx}
    \caption{Prompt template used to generate conversation rewrites. The \texttt{[NATURAL PROMPT]}, \texttt{[EMAIL]}, and \texttt{[REWRITE INSTRUCTION]} placeholders will be replaced by a given natural prompt, its raw generated email, and rewrite instruction.}
    \label{tab:conversation_rewrite_prompt}
\end{table*}

\begin{table*}[]
\small
    \centering
    \begin{tabularx}{\textwidth}{X}
        \toprule
        Task: Generate a rewrite instruction for a raw generated email based on the natural prompt. The instruction should guide the rewriting process to improve clarity, tone, structure, or other aspects. \\\\
        You will be given: \\
        1. Natural Prompt: The original prompt given to the LLM. \\
        2. Raw Generated Email: The initial email generated by the LLM based on the natural prompt. \\
        3. Demonstrations: Examples of existing emails and their rewrite instructions.\\\\
        Instructions: \\
        1. Read the `Natural Prompt' and `Raw Generated Email' carefully. \\
        2. Analyze the `Raw Generated Email' and identify areas for improvement based on the `Natural Prompt'. \\
        3. Refer to the provided `Demonstrations' to understand different types of rewrite instructions and their contexts. \\
        4. Generate a concise and clear rewrite instruction for the `Raw Generated Email'.\\\\
        Demonstrations: \\
        Example 1:\\
        - Email: \texttt{[EMAIL \#1]} \\
        - Instruction: \texttt{[REWRITE INSTRUCTION \#1]} \\
        ...\\
        Example N:\\
        - Email: \texttt{[EMAIL \#N]} \\
        - Instruction: \texttt{[REWRITE INSTRUCTION \#N]} \\\\
        Test example: \\
        1. Natural Prompt: \texttt{[NATURAL PROMPT]}\\
        2. Raw Generated Email: \texttt{[EMAIL]} \\
        3. Your Output (Rewrite Instruction):
        \\
    \bottomrule
    \end{tabularx}
    \caption{Prompt template used to generate raw conversation rewrite instructions. The \texttt{[NATURAL PROMPT]} and \texttt{[EMAIL]} placeholders will be replaced by a natural prompt and its generated raw email, and the \texttt{[EMAIL \#1 $\rightarrow$ N]} and \texttt{[REWRITE INSTRUCTION \#1 $\rightarrow$ N]} placeholders will be replaced by each of the N (email, instruction) pairs.}
    \label{tab:conversation_instruction_prompt}
\end{table*}

\noindent\textbf{Conversational rewrite prompt.}
Table~\ref{tab:conversation_rewrite_prompt} presents the prompt template for conversational rewrites, which is specifically designed to improve clarity, tone, and personalization in email responses. This prompt structure ensures that the model follows a controlled yet flexible rewriting process, preserving the intent of the original message while refining it according to a given instruction.
There are three key features of the prompt design:
(i) Explicit input segmentation: The prompt clearly separates the natural prompt, the raw generated email, and the rewrite instruction, ensuring that the model understands the original context before making modifications.
(ii) Focus on instruction adherence: Unlike factuality and stylistic rewrites, which prioritize correctness and rewording, this task emphasizes cohesively integrating the instruction into the existing email while maintaining internal consistency.
(iii) Preserving conversational flow: The instructions explicitly require the model to analyze the provided email and apply the requested changes while ensuring the rewritten email remains natural and engaging.

By structuring the prompt in a way that guides but does not over-restrict the model, this template ensures that email rewrites maintain fluency, correctness, and engagement while following specific improvement instructions. The step-by-step breakdown aids the model in handling more challenging contextual refinements, such as enhancing tone, incorporating enthusiasm, or making the message more concise.

\begin{table*}[htbp]
\small
\centering
\begin{tabularx}{\textwidth}{p{0.75\textwidth} X}
\toprule
\textbf{Raw email} & \textbf{Raw instruction} \\
\midrule
Dear James, 

I'd like to invite your to join ponding on [DATE]. Weather cold lately, recommend the wetsuit, gloves, woolly hat. Hope you can make it. Charles Campbell, the lifeguard, will also be joining us.

John 
& 
Rewrite using complete sentences. \\
\midrule
Sorry Alexander Thompson, my son, Frank Williams, can't attend, too young. Cancel registration \& refund. Sorry for any inconvenience. & Write in a formal tone.  \\
\midrule
Benjamin, check attached delivery notice for item.

Delivery person is Derek Johnson, arriving on December 14th between 10 am - 2 pm.

Need signature from someone over 18 to accept delivery.

Any questions, contact John Wilson. 
& 
Make it longer and more polite. \\
\midrule
Dear Anna Barret,

I am writing to request a change of class because I am currently in Kevin Smith's class and I think the teaching quality is very poor.

Sincerely,
Charles White & Mention that I don't learn well with Kevin's teaching style. \\
\midrule
Dear Connor,

I am writing today to request that you transfer my phone number from AT\&T to T-Mobile.

I have been an AT\&T customer for over 10 years and have been very happy with the service. However, I recently switched to T-Mobile because they have a better data plan for my needs. I would like to keep my AT\&T phone number so that my friends and family can still reach me.

Please let me know if you need any additional information from me. Thank you for your help.

Sincerely,

Katherine Walker 
& 
Makes it shorter. \\
\midrule 
Hi Avery,

I’m writing to request a refund for my valet reservation. Unfortunately, I’ve come down with a bad case of the flu and won’t be able to travel. I’m so sorry for the inconvenience.

I would really appreciate it if you could process the refund as soon as possible. Thank you for your understanding.

Sincerely,
Jerry Jones 
& 
This is too formal, makes it more casual. \\
\midrule 
Dear Liam,

I hope this email finds you well. I'm writing to see if you're available to cat-sit for James again. My family and I will be away on vacation from December 20th to January 2nd, and we would love it if you could take care of him while we're gone.

As you know, James is a very sweet and affectionate cat. He loves to be petted and cuddled, and he's always up for a game of fetch. He's also very good at entertaining himself, so you won't have to worry about him getting bored.

We would be happy to pay you the same rate as last time. Please let me know if you're available and if you have any questions.

Thank you, Fred 
& 
Don't mention that we will pay Liam.  \\
\midrule 
Hi Hailey,

I hope you're doing well. My name is Jerry Walker, and I'm currently in the process of purchasing a home in San Francisco. I was referred to you by George Wilson, who's a close friend of mine.

I recently had the home inspected, and the inspector identified a few electrical issues that I'd like to get fixed before moving in. Would you be able to provide me with a quote for the repairs? I'm planning to use the quote in my negotiations with the home seller. Thanks in advance for your help.

Best,
Jerry Walker 
& 
Make the second paragraph shorter. \\
\midrule 
Hi Ava,

Just following up on the insurance enrollment. I've been trying to reach Justin but no luck. Can u send me a written confirmation of my enrollment?

Thx.
Larry & Soften the tone to make this request more appealing. \\
\midrule 
Hi Ian, I will donate some SAS books, lmk if u want them.

Best,
Christopher & Add that the SAS books are "SAS Survival Handbook" and "The Little SAS Book". \\
\midrule 
Hi Andrew,

My gmail chat stopped working!!! I tried to log out/in, nothing works. Help!!!

Thx,
Ariana 
& 
Add structure and boilerplate sentences to this email to make it more professional. \\
\bottomrule
\end{tabularx}
\caption{Examples (demonstrations) of raw conversational rewrite instructions used in Table~\ref{tab:conversation_instruction_prompt}.}
\label{tab:raw_instruction_demonstrations}
\end{table*}

\noindent\textbf{Conversational rewrite instruction generation.} 
Table~\ref{tab:conversation_instruction_prompt} outlines the prompt template for generating rewrite instructions for conversational emails. This step is crucial in structuring the dataset, as it determines the type and specificity of modifications the model will learn to perform. Unlike factuality or stylistic rewrites, conversational rewrites often involve subtle refinements in tone, structure, and content balance.

Table~\ref{tab:raw_instruction_demonstrations} illustrates the diversity of generated instructions, ranging from simple grammatical refinements (e.g., “use complete sentences”) to substantial structural modifications (e.g., “add structure and boilerplate to make the email more professional”). The ability to generate rich, context-aware instructions ensures that conversational rewrites cover various real-world use cases, enhancing the model’s generalization ability.

\begin{table*}[]
\small
    \centering
    \begin{tabularx}{\textwidth}{X}
        \toprule
        Task: Modify the existing rewrite instruction to make it more specific. The modified instruction should provide clear and detailed guidance by specifying necessary entities, nouns, or details. \newline

        You will be given:\newline
        1. Natural Prompt: The original prompt given to the LLM.\newline
        2. Raw Generated Email: The initial email generated by the LLM based on the natural prompt.\newline
        3. Existing Instruction: The general rewrite instruction that needs to be refined.\newline
        4. Demonstrations: Examples showing how to modify general instructions to be more specific.\newline
        
        Instructions:\newline
        1. Read the 'Natural Prompt', 'Raw Generated Email', and 'Existing Instruction' carefully.\newline
        2. Refer to the provided 'Demonstrations' to understand how to modify an instruction to be more specific.\newline
        3. Modify the 'Existing Instruction' to make it more specific by including necessary entities, nouns, or details.\newline
        
        Demonstrations:\newline
        Example 1:\newline
        - Email: \texttt{[EMAIL \#1]} \newline
        - Original Instruction: \texttt{[ORIGINAL INSTRUCTION \#1]} \newline
        - Modified Instruction: \texttt{[MODIFIED INSTRUCTION \#1]} \newline
        ...\newline
        Example N:\newline
        - Email: \texttt{[EMAIL \#N]} \newline
        - Original Instruction: \texttt{[ORIGINAL INSTRUCTION \#N]} \newline
        - Modified Instruction: \texttt{[MODIFIED INSTRUCTION \#N]} \newline
        
        Test example: \newline
        1. Natural Prompt: \texttt{[NATURAL PROMPT]} \newline
        2. Raw Generated Email: \texttt{[EMAIL]} \newline
        3. Existing Instruction: \texttt{[ORIGINAL INSTRUCTION]} \newline
        4. Your Output (Modified Instruction):
        \\
    \bottomrule
    \end{tabularx}
    \caption{Prompt template used to specify conversational rewrite instructions with more details. The \texttt{[NATURAL PROMPT]}, \texttt{[EMAIL]}, and \texttt{[ORIGINAL INSTRUCTION]} placeholders will be replaced by a given natural prompt, its raw generated email, and original instruction. The \texttt{[EMAIL \#1 $\rightarrow$ N]}, \texttt{[ORIGINAL INSTRUCTION \#1 $\rightarrow$ N]}, and \texttt{[MODIFIED INSTRUCTION \#1 $\rightarrow$ N]} placeholders will be replaced by each of the N instruction refinement pairs.}
    \label{tab:conversation_instruction_specific_prompt}
\end{table*}

\begin{table*}[htbp]
\small
\centering
\begin{tabularx}{\textwidth}{p{0.52\textwidth} p{0.18\textwidth} X}
\toprule
\textbf{Raw email} & \textbf{Original instruction} & \textbf{Modified instruction} \\
\midrule
Dear [Recipient's name], I hope you have a great holiday season and enjoy your time with your family. & Make it more specific and personal. & Make it more specific and personal by mentioning the holiday plans you know about, such as their planned visit to Hawaii. \\
\midrule
Dear [Boss's name], Happy Mother's Day to you! I hope you have a wonderful day filled with love, laughter, and relaxation. I know that you are a great mother to your children, and I appreciate all that you do for them. You are always there for them, no matter what. You are a role model for them, and they are so lucky to have you. I also want to thank you for being a great boss. You are always supportive and understanding, and I appreciate your guidance and advice. I am lucky to have you as a mentor. I hope you have a very special Mother's Day! Sincerely, [Your name]" & Make it more personal and specific to my boss by mentioning her children's names or something special about her. & Make it more personal and specific to my boss by mentioning Eva, her daughter, and how she likes piggyback. \\
\midrule
Dear [Name], I hope this email finds you well. My name is [Your Name] and I am the event coordinator for [Event Name]. I am writing to you today to request a sponsorship from [Company Name]. [Event Name] is an annual event that raises money for [Charity Name]. Last year, we were able to raise over [Amount] for the charity. This year, we are hoping to raise even more money. We believe that [Company Name] would be a great fit for our event. Your company's products and services would be a great addition to our event and we believe that your customers would be interested in learning more about your company. In return for your sponsorship, we would be happy to provide you with a number of benefits, including: * Your logo would be prominently displayed on all promotional materials for the event. * You would have a booth at the event to promote your company. * You would be given the opportunity to speak at the event. * You would be invited to a VIP reception before the event. We believe that a sponsorship from [Company Name] would be a mutually beneficial relationship. We hope that you will consider our request. Thank you for your time and consideration. Sincerely, [Your Name] & Make the email more persuasive by highlighting the benefits of sponsorship for the company. & Make the email more persuasive by highlighting increased brand visibility, direct customer engagement at the event booth, and networking opportunities at the VIP reception. \\
\midrule
Dear [Patient Advocate Name], I am writing to you today to express my concern about the lack of informed consent in my recent medical treatment. On [date], I was admitted to [hospital name] for [procedure]. Prior to the procedure, I was asked to sign a consent form. However, I was not given any explanation of the procedure, the risks involved, or the alternatives to treatment. I was simply told that I needed to sign the form in order to receive treatment. I feel that I was not given the opportunity to make an informed decision about my treatment. I had no idea what I was consenting to, and I feel that I was taken advantage of. I am requesting that you investigate this matter and take steps to ensure that all patients are given the opportunity to make informed decisions about their treatment. Thank you for your time and consideration. Sincerely, [Your Name] & Add details about the procedure and the risks involved. & Add specific details about a knee replacement surgery and the potential risks of infection, blood clots, and damage to surrounding tissues. \\
\midrule
Dear Marketing Team, I hope this email finds you well. I'm writing to confirm that the account update has been completed. All of the necessary changes have been made to the account, and everything should be working properly. If you have any questions, please don't hesitate to contact me. Thank you for your cooperation. Sincerely, [Your Name] & Specify what account was updated and what changes were made. & Specify the account details such as client database, and describe the changes, like updated contact information, fixed email delivery issues, and added new segmentation tags. \\
\bottomrule
\end{tabularx}
\caption{Examples (demonstrations) of specified conversational rewrite instructions used in Table~\ref{tab:conversation_instruction_specific_prompt}.}
\label{tab:specify_instruction_demonstrations}
\end{table*}

\noindent\textbf{Conversational rewrite instruction refinement.} 
Table~\ref{tab:conversation_instruction_specific_prompt} presents the prompt template for refining generic rewrite instructions by making them more specific and actionable. This step is crucial in ensuring that rewrite instructions provide clear, detailed guidance, reducing ambiguity for the rewriting model. As seen in Table~\ref{tab:specify_instruction_demonstrations}, refining instructions significantly improves task clarity and execution. Instead of broad instructions (e.g., ``Make it more persuasive''), the specified versions provide concrete actionable changes (e.g., ``Highlight increased brand visibility, direct customer engagement, and networking opportunities''). This ensures that the rewriting model receives precise, contextually relevant directives, ultimately leading to higher-quality, more controlled rewrites.

\begin{table*}[]
\small
    \centering
    \begin{tabularx}{\textwidth}{X}
        \toprule
        Task: Modify the given rewrite instruction to make it more linguistically diverse and natural. The modified instruction should retain specificity and clarity while sounding more like human-asked natural language. \newline

        You will be given:\newline
        1. Natural Prompt: The original prompt given to the LLM.\newline
        2. Raw Generated Email: The initial email generated by the LLM based on the natural prompt.\newline
        3. Previous Instruction: The specific instruction generated previously that needs to be made more linguistically diverse and natural. \newline
        4. Demonstrations: Examples showing how to modify instructions to make them more natural and conversational. \newline
        
        Instructions:\newline
        1. Read the 'Natural Prompt', 'Raw Generated Email', and 'Existing Instruction' carefully.\newline
        2. Refer to the provided 'Demonstrations' to understand how to modify an instruction to be more specific.\newline
        3. Modify the 'Existing Instruction' to make it more specific by including necessary entities, nouns, or details.\newline
        
        Demonstrations:\newline
        Example 1:\newline
        - Email: \texttt{[EMAIL \#1]} \newline
        - Previous Instruction: \texttt{[ORIGINAL INSTRUCTION \#1]} \newline
        - Enhanced Instruction: \texttt{[MODIFIED INSTRUCTION \#1]} \newline
        ...\newline
        Example N:\newline
        - Email: \texttt{[EMAIL \#N]} \newline
        - Previous Instruction: \texttt{[ORIGINAL INSTRUCTION \#N]} \newline
        - Enhanced Instruction: \texttt{[MODIFIED INSTRUCTION \#N]} \newline
        
        Test example: \newline
        1. Natural Prompt: \texttt{[NATURAL PROMPT]} \newline
        2. Raw Generated Email: \texttt{[EMAIL]} \newline
        3. Previous Instruction: \texttt{[ORIGINAL INSTRUCTION]} \newline
        4. Your Output (Enhanced Instruction):
        \\
    \bottomrule
    \end{tabularx}
    \caption{Prompt template used to make conversational rewrite instructions more natural. The \texttt{[NATURAL PROMPT]}, \texttt{[EMAIL]}, and \texttt{[ORIGINAL INSTRUCTION]} placeholders will be replaced by a given natural prompt, its raw generated email, and previous instruction. The \texttt{[EMAIL \#1 $\rightarrow$ N]}, \texttt{[ORIGINAL INSTRUCTION \#1 $\rightarrow$ N]}, and \texttt{[MODIFIED INSTRUCTION \#1 $\rightarrow$ N]} placeholders will be replaced by each of the N instruction refinement pairs.}
    \label{tab:conversation_instruction_natural_prompt}
\end{table*}

\begin{table*}[htbp]
\small
\centering
\begin{tabularx}{\textwidth}{p{0.4\textwidth} p{0.3\textwidth} X}
\toprule
\textbf{Raw email} & \textbf{Original instruction} & \textbf{Modified instruction} \\
\midrule
Dear [Media Name], I hope this email finds you well. I'm writing to you today to invite you to a product update webinar on [date] at [time]. During this webinar, we'll be sharing the latest updates on our product, including new features, functionality, and improvements. We'll also be answering any questions you may have. We'd love for you to be able to attend. Please RSVP to [email address] by [date]. Thanks, [Your Name] & Make the email more specific by mentioning the product name and highlighting the key features of the update, such as the new AI-powered analytics tool and the redesigned user interface. Additionally, emphasize the benefits of these updates for media professionals, such as improved workflow efficiency and more insightful data analysis. & This email is too generic, specify the product name and highlight the key features of the update, such as the new AI-powered analytics tool and the redesigned user interface. Also emphasize the benefits of these updates for media professionals, such as improved workflow efficiency and more insightful data analysis. \\
\midrule 
Dear [Teacher's name], I hope this email finds you well. I'm writing to let you know that I recently got my dream job! I'm so excited to be starting this new chapter in my career. I wanted to thank you for all of your support and guidance over the years. You've helped me to become the confident and capable professional that I am today. I'll be sure to keep you updated on my progress. In the meantime, I wish you all the best. Sincerely, [Your name] & Make it more specific by mentioning the dream job as a [position name] at [company name], and how the teacher helped you achieve it by providing specific examples like [encouragement, career advice, mock interview practice, etc.]. & I just got a job as a graphic designer at disney, include a story about the time when Mr. Miller went through a bunch of van gogh paintings and how that really inspired me. \\
\midrule 
Dear Mom, I hope this email finds you well. I just wanted to take a moment to thank you for coming to my birthday party last night. I had so much fun celebrating with you and all of my friends and family. It meant the world to me to have you there. I know you were busy, but I'm so glad you made the time. I hope you had a good time as well. I know I did! Love, [Your name] & Make it more personal and heartfelt by mentioning specific moments you enjoyed with your mom during the party, like dancing together or her funny stories. You can also express your love and appreciation for her presence and support. & Make the email more heartfelt, including reference to the times we were dancing together and her funny stories. \\
\bottomrule
\end{tabularx}
\caption{Examples (demonstrations) of naturalized conversational rewrite instructions used in Table~\ref{tab:conversation_instruction_natural_prompt}.}
\label{tab:naturalize_instruction_demonstrations}
\end{table*}

Table~\ref{tab:conversation_instruction_natural_prompt} presents the prompt template for making rewrite instructions more natural and linguistically diverse. This step is essential in ensuring that rewrite instructions feel human-like, conversational, and engaging, while still preserving clarity and specificity. Table~\ref{tab:naturalize_instruction_demonstrations} showcases how structured rewrite instructions are transformed into more intuitive, engaging requests. Instead of formal, mechanical directives (e.g., ``Make the email more specific by mentioning the product name and key features.''), the modified instructions use more natural phrasing (e.g., ``This email is too generic—specify the product name and key features, and highlight why these updates matter for media professionals.'').

\section{AutoRater Prompting}
\label{sec:evaluation_prompts}

The LLM-as-a-judge framework systematically evaluates rewrites across agreement, coherence, and pairwise comparisons (AutoSxS) to ensure fine-grained, objective assessments. Below, we describes its design rationales.

\noindent\textbf{Agreement evaluation across tasks.}
Agreement evaluation measures how well the rewritten response adheres to the given instruction across different tasks:
\begin{itemize}[nosep,leftmargin=*]
    \item \textbf{Factuality rewrite}: Table~\ref{tab:factuality_rewrite_agreement_prompt} illustrates a span-based approach, checking whether all identified factual inaccuracies in the initial response have been corrected. Each non-factual span is individually assessed, ensuring precise, granular feedback.
    \item \textbf{Stylistic rewrite}: Table~\ref{tab:style_rewrite_agreement_prompt} deconstructs the instruction into multiple transformation requirements (e.g., formalization + conciseness) and evaluate whether each aspect is incorporated while preserving meaning.
    \item \textbf{Conversational rewrite}: Table~\ref{tab:conversation_rewrite_agreement_prompt} introduces a context-aware evaluation, ensuring that modifications align with the natural prompt while accurately implementing tone, structure, or engagement-related changes.
\end{itemize}

\begin{table*}[]
\small
    \centering
    \begin{tabularx}{\textwidth}{X}
        \toprule
        Task: Analyze a revised text passage for factual accuracy by calculating the percentage of replacements correctly incorporated into the revised version. \newline

        You will be given: \newline
        1. Query: The user query. \newline
        2. Initial Response: The original text generated by an LLM to answer the user query. \newline
        3. Non-Factual Spans and Replacements: A list of text segments from the initial response that were potentially incorrect, along with their intended factual replacements. \newline
        4. Rewritten Response: A revised text to answer the user query where the goal was to replace the non-factual spans with their factual counterparts while making minimal edits. \newline
        
        Instructions: \newline
        1. Carefully compare the `Non-Factual Spans and Replacements' and the `Rewritten Response'. Note whether sentence structure changes were necessary to integrate the replacements while striving for accuracy. \newline
        2. Check Accuracy: Meticulously examine the `Rewritten Response'. Your focus is to determine if ALL intended factual replacements are correctly incorporated. \newline
        3. Calculate Percentage: Calculate the following: \newline
        Total Number of Replacements: The number of items in the `Non-Factual Spans and Replacements' list. \newline
        Correct Replacements: The number of replacements accurately integrated into the `Rewritten Response'. \newline
        Percentage: (Correct Replacements / Total Number of Replacements) * 100 \newline
        4. Output Percentage. \newline
        
        Test example: \newline
        1. Query: \texttt{[QUERY]} \newline
        2. Initial Response: \texttt{[INITIAL RESPONSE]} \newline
        3. Non-Factual Spans and Replacements: \newline  
        Span 1: \newline
        - Span: \texttt{[SPAN \#1]} \newline
        - Revision: \texttt{[REVISION \#1]} \newline
        ... \newline
        Span N: \newline
        - Span: \texttt{[SPAN \#N]} \newline
        - Revision: \texttt{[REVISION \#N]} \newline
        4. Rewritten Response: \texttt{[REWRITTEN RESPONSE]} \newline
        5. Your Output: 
        \\
    \bottomrule
    \end{tabularx}
    \caption{Prompt template used to generate factuality rewrite agreement score. The \texttt{[QUERY]}, \texttt{[INITIAL RESPONSE]}, and \texttt{[REWRITTEN RESPONSE]} placeholders will be replaced by a given user query, its initial response, and the rewritten response, and the \texttt{[SPAN \#1 $\rightarrow$ N]} and \texttt{[REVISION \#1 $\rightarrow$ N]} placeholders will be replaced by each of the N spans and revisions from the SAFE~\cite{wei2024long} outputs.}
    \label{tab:factuality_rewrite_agreement_prompt}
\end{table*}

\begin{table*}[]
\small
    \centering
    \begin{tabularx}{\textwidth}{X}
        \toprule
        Task: Analyze a revised text passage for agreement with the given rewrite instruction. \newline

        You will be given: \newline
        1. Source: The original text from a public corpus. \newline
        2. Instruction: Open-ended rewrite requirements, such as formalize, paraphrase, shorten, elaborate, etc. \newline
        3. Rewritten text: The revised text that has been altered according to the instruction's requirements. \newline
        
        Instructions: \newline
        1. Carefully compare the 'source' and the 'rewritten text' based on the 'instruction'. Note whether the structure, tone, and content changes were necessary to integrate the instruction's requirements while maintaining coherence and consistency. \newline
        2. Check Agreement: Meticulously examine the 'rewritten text'. Your focus is to determine if ALL requirements mentioned in the 'instruction' are correctly and effectively incorporated. \newline
        3. Calculate Percentage: Calculate the following: \newline
        Total Number of Requirements: The number of distinct requirements mentioned in the 'instruction'. \newline
        Correct Requirements: The number of requirements accurately integrated into the 'rewritten text'. \newline
        Percentage: (Correct Requirements / Total Number of Requirements) * 100 \newline
        4. Output Percentage. \newline
        
        
        Test example: \newline
        1. Source: \texttt{[SOURCE]} \newline
        2. Instruction: \texttt{[INSTRUCTION]} \newline
        3. Rewritten text: \texttt{[REWRITTEN TEXT]} \newline
        4. Your Output: 
        \\
    \bottomrule
    \end{tabularx}
    \caption{Prompt template used to generate stylistic rewrite agreement score. The \texttt{[SOURCE]}, \texttt{[INSTRUCTION]}, and \texttt{[REWRITTEN TEXT]} placeholders will be replaced by a given source text, rewrite instruction, and the rewritten text.}
    \label{tab:style_rewrite_agreement_prompt}
\end{table*}

\begin{table*}[]
\small
    \centering
    \begin{tabularx}{\textwidth}{X}
        \toprule
        Task: Analyze a revised email for agreement with the rewrite instruction.\newline

        You will be given this XML input:\newline
        <natural\_prompt>The original prompt given to the LLM to generate the initial email.</natural\_prompt>\newline
        <raw\_generated\_email>The initial email generated by the LLM based on the natural prompt.</raw\_generated\_email>\newline
        <rewrite\_instruction>The instruction provided for rewriting the initial email.</rewrite\_instruction>\newline
        <rewritten\_email>The revised email that has been altered according to the rewrite instruction.</rewritten\_email>\newline
        
        Instructions:\newline
        1. Carefully compare the 'Raw Generated Email' and the 'Rewritten Email' based on the 'Rewrite Instruction'. Note whether the structure, tone, and content changes were necessary to integrate the instruction's requirements while maintaining coherence and consistency.\newline
        2. Check Agreement: Meticulously examine the 'Rewritten Email'. Your focus is to determine if ALL requirements mentioned in the 'Rewrite Instruction' are correctly and effectively incorporated.\newline
        3. Calculate Percentage: Calculate the following:\newline
        Total Number of Requirements: The number of distinct requirements mentioned in the 'Rewrite Instruction'.\newline
        Correct Requirements: The number of requirements accurately integrated into the 'Rewritten Email'.\newline
        Percentage: (Correct Requirements / Total Number of Requirements) * 100\newline
        4. Output Percentage.\newline
        
        You will output in XML form:\newline
        <output\_explanation>Explain your reasoning for each requirement, indicating whether it's satisfied, and why.</output\_explanation>\newline
        <output\_percentage>Provide the final accuracy percentage. Do not include anything besides a percentage.</output\_percentage>\newline
        
        Begin!\newline
        
        <natural\_prompt>\texttt{[NATURAL PROMPT]}</natural\_prompt>\newline
        <raw\_generated\_email>\texttt{[EMAIL]}</raw\_generated\_email>\newline
        <rewrite\_instruction>\texttt{[REWRITE INSTRUCTION]}</rewrite\_instruction>\newline
        <rewritten\_email>\texttt{REWRITTEN EMAIL}</rewritten\_email>\newline
        
        Your XML output with explanation and percentage:\newline
        <output\_explanation>
        \\
    \bottomrule
    \end{tabularx}
    \caption{Prompt template used to generate conversational rewrite agreement score. The \texttt{[NATURAL PROMPT]}, \texttt{[EMAIL]}, \texttt{[REWRITE INSTRUCTION]}, and \texttt{REWRITTEN EMAIL} placeholders will be replaced by a given natural prompt, its raw generated email, rewrite instruction, and rewritten email.}
    \label{tab:conversation_rewrite_agreement_prompt}
\end{table*}

\noindent\textbf{Coherence evaluation.}
Coherence evaluation in AutoRater is designed to assess whether a rewritten response maintains internal logical consistency while ensuring fluency and readability. Unlike agreement evaluation, which focuses on compliance with rewrite instructions, coherence evaluation directly measures whether a revised response contains contradictions, logical inconsistencies, or abrupt structural breaks.
The AutoRater employs a structured prompt (Table~\ref{tab:coherence_prompt}) that systematically evaluates coherence by requiring explicit yes/no judgments alongside explanatory reasoning. This approach enhances interpretability and ensures reliable scoring.
(i) Binary Coherence Decision: Each rewritten response is judged as either internally consistent ("YES") or inconsistent ("NO"), ensuring clarity in evaluation.
(ii) Justification for Each Judgment: The prompt mandates a reason for the decision, encouraging the model to articulate why a response is or is not coherent.
(iii) Few-Shot Learning with Examples: The prompt provides multiple illustrative cases, demonstrating correct coherence judgments across different scenarios, including numerical inconsistencies, logical contradictions, and unsupported claims.

\begin{table*}[]
\small
    \centering
    \begin{tabularx}{\textwidth}{X}
        \toprule
        Your task is to evaluate whether or not a piece of text is internally consistent. \newline

        Please provide your answer in the following format: \newline
        <text>[text to evaluate for internal consistency]</text> \newline
        <answer>[YES or NO]</answer> \newline
        <reason>[reason for giving the answer above]</reason> \newline
        
        Example \#1: \newline
        <text>Paul has 5 daughters named Ava, Brittney, and Claire.</text> \newline
        <answer>NO</answer> \newline
        <reason>The response states that Paul has 5 daughters but only names 3 of them.</reason> \newline
        
        Example \#2: \newline
        <text>IPhones are better than Samsung. The reason is because Samsung has a shorter battery life.</text> \newline
        <answer>YES</answer> \newline
        <reason>Everything is internally consistent.</reason> \newline
        
        Example \#3: \newline
        <text>While Bob Ross and Annettte Kowalski's exact age difference is not publicly known, they likely had several years age difference. \newline
        
        Here's what we found online: \newline
        * Bob Ross was born on October 29, 1942 and died at age 52 on July 4, 1995. \newline
        * Annette Kowalski was born on January 26, 1936. \newline
        </text> \newline
        <answer>NO</answer> \newline
        <reason>The beginning says that the age difference is not publicly known, but later the text says that both birth dates were found online.</reason> \newline
        
        Question: \newline
        <text>\texttt{[REWRITTEN RESPONSE]}</text> \newline
        <answer>
        \\
    \bottomrule
    \end{tabularx}
    \caption{Prompt template used to generate rewrite coherence score. The \texttt{[REWRITTEN RESPONSE]} placeholder will be replaced by a rewritten response.}
    \label{tab:coherence_prompt}
\end{table*}

\noindent\textbf{AutoSxS evaluation.}
AutoSxS is designed to directly compare rewritten responses by evaluating how well they satisfy a given rewrite instruction. Unlike individual scoring metrics (e.g., agreement, coherence), AutoSxS provides pairwise judgments, making it a more fine-grained and human-aligned evaluation method.
As shown in Table~\ref{tab:autosxs_prompt}, the AutoSxS framework follows a structured decision-making process:
(i) The model is presented with two rewritten responses (A, B) for the same raw email and rewrite instruction.
(ii) It must analyze both responses and select the one that better satisfies the rewrite instruction (or declare them as equal).
(iii) It provides explicit reasoning for its choice and assigns numerical scores (0–1) to each response, ensuring that the relative ranking is interpretable.
To enhance reliability, the prompt includes:
(i) Clear task instructions emphasizing rewrite alignment.
(ii) Demonstrations with diverse examples covering different types of modifications (removal, addition, tone change, etc.).
(iii) Structured output, ensuring consistency in comparisons.

Table~\ref{tab:autosxs_examples} presents two examples illustrating AutoSxS decisions.
In Example 1, Response A is rated 0.91, significantly higher than Response B (0.58), as it strictly adheres to the instruction of removing unnecessary details.
In Example 2, Response B is preferred (0.83 vs. 0.76) for expanding on the religious holiday's significance in a more formal and comprehensive way.
These results demonstrate AutoSxS's ability to:
(i) Identify minor instruction violations (e.g., residual details in Response B of Example 1).
(ii) Capture nuanced preferences in tone and detail level (e.g., preference for a more formal explanation in Example 2).
(iii) Provide interpretable ranking beyond binary correctness.
By leveraging structured evaluation, explicit reasoning, and numerical scoring, AutoSxS enhances automated rewrite assessment, ensuring more human-aligned and context-aware comparisons across rewriting tasks.

\begin{table*}[]
\small
    \centering
    \begin{tabularx}{\textwidth}{X}
        \toprule
        Task: Judge which response better satisfies the rewrite instruction, choose among two choices: (A), (B) or same, and rate each response a score between 0 and 1. \newline
        Instructions: \newline
        1. Carefully analyze the `Response A' and `Response B' based on the `Instruction'. \newline
        2. Provide an explanation comparing the two responses based on the instruction. \newline
        3. Make your choice between `A', `B', or `same'. \newline
        4. Provide your score for each response (A, B) based on how well they satisfy the rewrite instruction. \newline
        
        Demonstrations: \newline
        Example \#1 \newline
        \hspace*{10pt} <original\_email>\texttt{[EMAIL \#1]}</original\_email> \newline
        \hspace*{10pt} <instruction>\texttt{[REWRITE INSTRUCTION \#1]}</instruction> \newline
        \hspace*{10pt} <response\_A>\texttt{[RESPONSE A \#1]}</response\_A> \newline
        \hspace*{10pt} <response\_B>\texttt{[RESPONSE B \#1]}</response\_B> \newline
        \hspace*{10pt} <choice>\texttt{[CHOICE \#1]}</choice> \newline
        \hspace*{10pt} <explanation>\texttt{[REASON \#1]}</explanation> \newline
        \hspace*{10pt} <score>\texttt{[SCORE TUPLE \#1]}</score> \newline
        ... \newline
        Example \#N \newline
        \hspace*{10pt} <original\_email>\texttt{[EMAIL \#N]}</original\_email> \newline
        \hspace*{10pt} <instruction>\texttt{[REWRITE INSTRUCTION \#N]}</instruction> \newline
        \hspace*{10pt} <response\_A>\texttt{[RESPONSE A \#N]}</response\_A> \newline
        \hspace*{10pt} <response\_B>\texttt{[RESPONSE B \#N]}</response\_B> \newline
        \hspace*{10pt} <choice>\texttt{[CHOICE \#N]}</choice> \newline
        \hspace*{10pt} <explanation>\texttt{[REASON \#N]}</explanation> \newline
        \hspace*{10pt} <score>\texttt{[SCORE TUPLE \#N]}</score> \newline
        
        You will output in XML format: \newline
        <choice>Please choose among three choices: (A), (B) or same. Do not include anything besides a choice.</choice> \newline
        <explanation>Explain the reason of your choice.</explanation> \newline
        <score>A tuple of two scores between 0 and 1, rating how well each response satisfies the rewrite instruction.</score> \newline
        
        Begin! \newline
        <original\_email>\texttt{[EMAIL]} </original\_email> \newline
        <instruction>\texttt{[REWRITE INSTRUCTION]}</instruction> \newline
        <response\_A>\texttt{[RESPONSE A]}</response\_A> \newline
        <response\_B>\texttt{[RESPONSE B]}</response\_B> \newline
        
        Your XML output with choice, explanation, and score: \newline
        <choice>
        \\
    \bottomrule
    \end{tabularx}
    \caption{Prompt template used to generate AutoSxS scores of two rewritten responses $\texttt{[RESPONSE A]}$, $\texttt{[RESPONSE B]}$ regarding the same raw generated email $\texttt{[EMAIL]}$ and instruction $\texttt{[REWRITE INSTRUCTION]}$.}
    \label{tab:autosxs_prompt}
\end{table*}

\begin{table*}[]
\small
\centering
\begin{tabularx}{\textwidth}{p{0.48\textwidth} p{0.48\textwidth}}
\toprule
\multicolumn{2}{p{0.98\textwidth}}{\textbf{Raw email:} Dear [Name], \newline
I am writing to cancel the interview with the shareholders that was scheduled for [date]. I am very sorry for any inconvenience this may cause. I have a personal emergency that I need to attend to, and I will not be able to make it to the interview. \newline
I have already contacted [Name], and they have agreed to reschedule the interview for [date]. \newline
I apologize again for any inconvenience this may cause. Thank you for your understanding. \newline
Sincerely, [Your Name]}
 \\\\

\multicolumn{2}{p{0.98\textwidth}}{\textbf{Instruction:} Remove the sentences mentioning the personal emergency and contacting [Name]. Instead, state that the interview is being canceled due to unforeseen circumstances.
} \\\\

\textbf{Response (A):} Dear [Name], \newline
I am writing to inform you that, due to unforeseen circumstances, I must cancel the interview with the shareholders scheduled for [date]. Please accept my sincerest apologies for any inconvenience this may cause. \newline
I understand this may be disappointing news, and I appreciate your understanding during this time. If necessary, we can discuss rescheduling at a later date. Thank you for your continued support. \newline
Sincerely, [Your Name]
 &
\textbf{Response (B):} Dear [Name], \newline
I am writing to cancel the interview with the shareholders scheduled for [date] due to unforeseen circumstances. I apologize for any inconvenience this may cause. Unfortunately, I won't be able to attend as planned, and I will follow up with [Name] regarding rescheduling. \newline
Thank you for your understanding and patience. \newline
Best regards, [Your Name] \\\\

\textbf{Score (A):} 0.91 & \textbf{Score (B):} 0.58 \\\\

\textbf{Choice:} (A) & \\\\

\multicolumn{2}{p{0.98\textwidth}}{\textbf{Reasoning:} Response (A) directly follows the instruction by focusing entirely on the unforeseen circumstances and does not mention contacting [Name] or personal emergencies. It also offers to discuss rescheduling in a more formal tone. Response (B) still includes a mention of following up with [Name], which was not in the instruction, and has a more casual tone, which might make it slightly less aligned with the rewrite instruction.} \\

\midrule

\multicolumn{2}{p{0.98\textwidth}}{\textbf{Raw email:} Dear [Recipient Name], \newline 
I am writing to request leave for [number of days] days on [date] and [date] for [religious holiday]. \newline 
I understand that this is a busy time for the team, but I would be grateful if you could approve my request. I have already completed all of my tasks for the week, and I have made arrangements for [coworker's name] to cover my duties while I am away. Thank you for your understanding. \newline 
Sincerely, [Your Name]} \\\\

\multicolumn{2}{p{0.98\textwidth}}{\textbf{Instruction:} Please add a sentence or two about the significance of the religious holiday and how you plan to observe it.} \\\\

\textbf{Response (A):} Dear [Recipient Name], \newline
I am writing to request leave for [number of days] days on [date] and [date] for [religious holiday]. This is a special time for me as it allows me to connect with my family and our traditions. We plan to celebrate with a gathering and special prayers. \newline
I have ensured that all my tasks are completed and that [coworker's name] will handle any urgent matters in my absence. I hope for your approval of my leave request. Thank you for your understanding. \newline
Best regards, [Your Name]
&
\textbf{Response (B):} Dear [Recipient Name], \newline
I am writing to request leave for [number of days] days on [date] and [date] for [religious holiday]. This holiday holds significant spiritual importance for me and my family. We plan to observe it by participating in religious ceremonies and spending quality time together. \newline
I realize that my absence may impact the team's workload, and I have made arrangements for [coworker's name] to cover my duties during this period. I would greatly appreciate your approval for this leave.
Thank you for your consideration. \newline
Sincerely, [Your Name]
 \\\\

\textbf{Score (A):} 0.76 & \textbf{Score (B):} 0.83 \\\\

\textbf{Choice:} (B) & \\\\

\multicolumn{2}{p{0.98\textwidth}}{\textbf{Reasoning:} Both responses add information about the significance of the religious holiday, but they do so in different ways. Response (A) emphasizes the family connection and traditional practices, while Response (B) focuses on the spiritual importance and includes participation in ceremonies. Both are valid, but Response (B) might be seen as slightly more formal and aligned with the instruction to expand on the holiday's significance.} \\
\bottomrule
\end{tabularx}
\caption{Two examples used for prompting AutoSxS evaluations in Table~\ref{tab:autosxs_prompt}.}
\label{tab:autosxs_examples}
\end{table*}

\end{document}